\documentclass[10pt,journal,compsoc]{IEEEtran}

\ifCLASSOPTIONcompsoc
  \usepackage[nocompress]{cite}
  \usepackage[caption=false,font=footnotesize,labelfont=sf,textfont=sf]{subfig}
\else
  \usepackage{cite}
  \usepackage[caption=false,font=footnotesize]{subfig}
\fi
\ifCLASSINFOpdf
  \usepackage[pdftex]{graphicx}
  \DeclareGraphicsExtensions{.pdf,.jpeg,.png}
\else
  \usepackage[dvips]{graphicx}
  \DeclareGraphicsExtensions{.eps}
\fi
\usepackage{amsmath,amssymb,amsfonts,dsfont,bm,mathrsfs}
\usepackage{algorithmic}
\usepackage{booktabs,multirow,adjustbox}
\usepackage{float}
\usepackage[table]{xcolor}
\usepackage{collcell}
\usepackage{xifthen}
\usepackage{pgf}
\usepackage[pagebackref=false,breaklinks=true,colorlinks=true,citecolor=blue,bookmarks=false]{hyperref}
\def\ColorMode{hsb}
\newcommand{\ColCell}[1]{
  \ifthenelse{\isempty{#1}}{}{
    \pgfmathparse{#1<2?1:0}                          
      \ifnum\pgfmathresult=0\relax\color{white}\fi
    \pgfmathsetmacro\compA{0}                        
    \pgfmathsetmacro\compB{(abs(#1)==1?0:abs(#1))/1} 
    \pgfmathsetmacro\compC{1}                        
    \edef\x{\noexpand\centering\noexpand\cellcolor[\ColorMode]{\compA,\compB,\compC}}\x #1
  }
}
\newcolumntype{C}[1]{>{\collectcell\ColCell}m{#1}<{\endcollectcell}}  
\newcommand{\rot}{\rotatebox{45}}
\newcommand{\NumAttributes}{5}

\newcommand{\R}{\mathbb{R}}      
\newcommand{\E}{\mathbb{E}}      
\newcommand{\Prob}{{\rm P}}      
\newcommand{\Norm}{\mathcal{N}}  
\newcommand{\0}{{\rm\bf 0}}      
\newcommand{\I}{{\rm\bf I}}      
\newcommand{\z}{{\rm\bf z}}      
\newcommand{\Z}{\mathcal{Z}}     
\newcommand{\w}{{\rm\bf w}}      
\newcommand{\W}{\mathcal{W}}     
\newcommand{\x}{{\rm\bf x}}      
\newcommand{\X}{\mathcal{X}}     
\newcommand{\s}{{\rm\bf s}}      
\renewcommand{\S}{\mathcal{S}}   
\newcommand{\n}{{\rm\bf n}}      
\newcommand{\N}{{\rm\bf N}}      
\newcommand{\dis}{{\rm d}}       

\begin{document}

\newcommand{\titlename}{InterFaceGAN: Interpreting the Disentangled Face Representation Learned by GANs}
\title{\titlename}

\author{
  Yujun Shen,
  Ceyuan Yang,
  Xiaoou Tang, \IEEEmembership{Fellow, IEEE,} and
  Bolei Zhou, \IEEEmembership{Member, IEEE}
  \IEEEcompsocitemizethanks{
    \IEEEcompsocthanksitem Y. Shen, C. Yang, X. Tang, and B. Zhou are with
    the Department of Information Engineering, the Chinese University of Hong Kong, Hong Kong SAR, China.\protect\\
    E-mail: \{sy116, yc019, xtang, bzhou\}@ie.cuhk.edu.hk
  }%
}


\IEEEtitleabstractindextext{

\begin{abstract}
Although Generative Adversarial Networks (GANs) have made significant progress in face synthesis, there lacks enough understanding of what GANs have learned in the latent representation to map a random code to a photo-realistic image.
In this work, we propose a framework called InterFaceGAN to interpret the disentangled face representation learned by the state-of-the-art GAN models and study the properties of the facial semantics encoded in the latent space.
We first find that GANs learn various semantics in some linear subspaces of the latent space.
After identifying these subspaces, we can realistically manipulate the corresponding facial attributes without retraining the model.
We then conduct a detailed study on the correlation between different semantics and manage to better disentangle them via subspace projection, resulting in more precise control of the attribute manipulation.
Besides manipulating the gender, age, expression, and presence of eyeglasses, we can even alter the face pose and fix the artifacts accidentally made by GANs.
Furthermore, we perform an in-depth face identity analysis and a layer-wise analysis to evaluate the editing results quantitatively.
Finally, we apply our approach to real face editing by employing GAN inversion approaches and explicitly training feed-forward models based on the synthetic data established by InterFaceGAN.
Extensive experimental results suggest that learning to synthesize faces spontaneously brings a disentangled and controllable face representation.
\thanks{Code and models are available at \protect\url{https://genforce.github.io/interfacegan/}.}
\end{abstract}


}

\maketitle
\IEEEdisplaynontitleabstractindextext
\IEEEpeerreviewmaketitle

\begin{figure*}[t]
  \centering
  \includegraphics[width=1.0\linewidth]{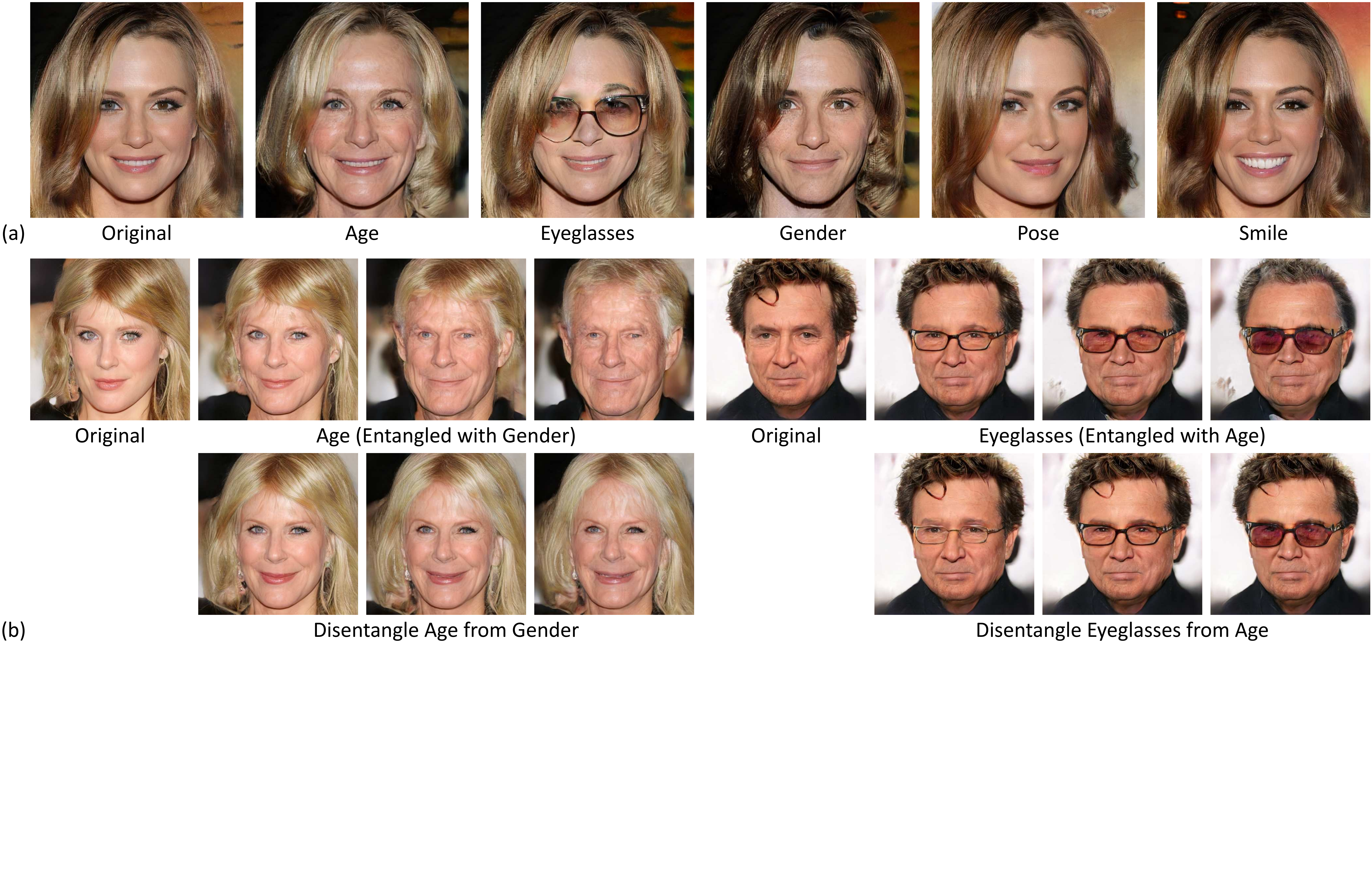}
  \vspace{-20pt}
  \caption{
   ~(a) \textbf{Manipulating various facial attributes} through varying the latent codes of a well-trained GAN model.
   ~(b) \textbf{Conditional manipulation} results using InterFaceGAN, where we can better disentangle the correlated attributes (top row) and achieve more precise control of the facial attributes (bottom row).
    All results are synthesized by PGGAN \cite{pggan}.
  }
  \label{fig:teaser}
  \vspace{0pt}
\end{figure*}

\IEEEraisesectionheading{\section{Introduction}\label{sec:introduction}}
\IEEEPARstart{R}{ecent} years have witnessed the great success of Generative Adversarial Networks (GANs) \cite{gan} in high-fidelity face synthesis \cite{pggan,stylegan,stylegan2}.
Based on adversarial training, GANs learn to map a random distribution to the real data observation and then produce photo-realistic images from randomly sampled latent codes.

Despite the appealing synthesis quality, it remains much less explored about what knowledge GANs learn in the latent representation and how we can reuse such knowledge to control the generation process.
For example, given a latent code, how does GAN determine the attributes of the output face, \emph{e.g.}, an elder man or a young woman?
How are different attributes organized in the latent space?
Can we manipulate the attributes of the synthesized face as we want?
How does the attribute manipulation affect the face identity?
Can we apply a well-trained GAN model for real image editing?

To answer these questions, we propose a novel framework, termed as \emph{InterFaceGAN}, to \emph{Inter}pret the \emph{Face} representation learned by \emph{GAN}s.
We employ some off-the-shelf classifiers to predict semantic scores for synthesized images.
In this way, we can bridge the latent space and the semantic space and further utilize such a connection for representation analysis.
In particular, we analyze how an individual semantic is encoded in the latent space both theoretically and empirically.
It turns out that a true-or-false attribute aligns with a linear subspace of the latent space.
Based on this discovery, we study the entanglement between different semantics emerging in the latent representation and manage to disentangle them via subspace projection.

After identifying the latent semantics, InterFaceGAN proposes a simple yet effective pipeline for face editing.
By linearly varying the latent code, we can manipulate the gender, age, expression, presence of eyeglasses, and even face pose of the synthesized image, as shown in Figure~\ref{fig:teaser}~(a).
Moreover, thanks to our disentanglement analysis, we propose conditional manipulation to alter one attribute without affecting others, as shown in Figure~\ref{fig:teaser}~(b).
More importantly, InterFaceGAN controls the face semantics by reusing the GAN knowledge \emph{without} retraining the model.

To better interpret the representation learned by GANs, we conduct a thorough analysis of the editing results made by InterFaceGAN.
First, we compare the semantic scores before and after the manipulation to quantitatively evaluate the identified semantics.
Then, we make a layer-wise analysis on StyleGAN, whose generator is with per-layer stochasticity \cite{stylegan}, to explore how semantics originate from the latent representation layer by layer.
Finally, considering the importance of the identity information for faces, we make in-depth identity analysis to see how identity is preserved in the manipulation process as well as how identity is sensitive to different facial attributes.

Applying a GAN model to real image editing is another important issue since GANs commonly lack the inference ability.
In this work, we apply two approaches to extend InterFaceGAN for real face manipulation.
One works with GAN inversion, which can invert a target image back to a latent code.
The other uses InterFaceGAN to build a dataset that contains the pairs of synthetic images before and after manipulation and then train a pixel-to-pixel model \cite{pix2pixhd} on this synthetic dataset.
We compare these approaches and evaluate their strengths and weaknesses.

The preliminary result of this work is published at~\cite{interfacegan}.
Compared to the conference paper, we include the following new contents:
(i) a detailed analysis of the face representation learned by StyleGAN \cite{stylegan} and its comparison to PGGAN \cite{pggan};
(ii) a comparison between the entanglement of identified latent semantics and the attribute distribution of training data, which sheds light on how GANs learn to encode various semantics in the training process;
(iii) quantitative evaluation of the editing results achieved by InterFaceGAN;
(iv) layer-wise analysis of the per-layer representation learned by StyleGAN \cite{stylegan};
(v) identity analysis of the manipulated images;
(vi) a new method for real face editing, which is to train feed-forward models on the synthetic data collected by InterFaceGAN.

\section{Related Work}\label{sec:related-work}

\vspace{2pt}\noindent\textbf{Generative Adversarial Networks.}
Due to the great potential of GAN \cite{gan} in producing photo-realistic images, it has been widely applied to image editing \cite{lample2017fader,bau2019semantic}, super-resolution \cite{ledig2017photo,wang2018esrgan}, image inpainting \cite{yeh2017semantic,yu2019free}, video synthesis \cite{wang2018video,wang2019few}, \emph{etc}.
Many attempts have been made to improve the synthesis quality and training stability of GANs \cite{wgan,wgan_gp,began,sngan,sagan,pggan,biggan,stylegan,stylegan2}.
Despite this tremendous success, little work has been done on understanding how GANs learn to connect the latent representation with the semantics in the real visual world.

\vspace{2pt}\noindent\textbf{Study on Latent Space of GANs.}
Latent space of GANs is generally treated as Riemannian manifold \cite{gan_metrics,latent_oddity,kuhnel2018latent}.
Prior work focused on exploring how to make the output image vary smoothly from one synthesis to another through interpolation in the latent space \cite{feature_based_metrics,riemannian_geometry}.
Bojanowski \emph{et al.} \cite{glo} optimized the generator and latent code simultaneously to learn a better representation.
However, the studies on how a well-trained GAN can encode different semantics in the latent space and reuse this semantic knowledge to control the generation process are still missing.
Bau \emph{et al.} \cite{bau2019gandissect} found that some units from intermediate layers of the GAN generator are specialized to synthesize certain visual concepts, such as sofa and TV for living room synthesis.
Some work \cite{dcgan,feature_interpolation} observed the vector arithmetic property in the latent space.
This work provides a detailed analysis of how semantics are encoded in the face representation learned by GANs from the perspective of a single semantic and the disentanglement of multiple semantics.
Some concurrent work also explores the latent semantics in GANs:
Goetschalckx \emph{et al.} \cite{goetschalckx2019ganalyze} improves the memorability of the output image.
Jahanian \emph{et al.} \cite{gansteerability} studies the steerability of GANs concerning camera motion and image color tone.
Yang \emph{et al.} \cite{yang2019semantic} observes the semantic hierarchy emerging in the scene synthesis models.
Differently, we focus on interpreting the face representation by theoretically and empirically studying how various semantics originate from and are organized in the latent space.
We further extend our method to real image editing.

\vspace{2pt}\noindent\textbf{Semantic Face Editing with GANs.}
Semantic face editing aims at manipulating the facial attributes of a given image.
Compared to unconditional GANs, which can generate images arbitrarily, semantic editing expects the model to change the target attribute yet maintain other information of the input face.
To achieve this goal, existing methods require carefully designed loss functions \cite{acgan,infogan,drgan}, introduction of additional attribute labels \cite{lample2017fader,ffgan,opensetgan,elegant,facefeatgan}, or special architectures \cite{sdgan,faceidgan} to train new models.
Unlike previous learning-based methods, this work explores the interpretable semantics inside the latent space of \emph{fixed} GAN models.
By reusing the semantic knowledge spontaneously learned by GANs, we can unleash its manipulation capability and \emph{turn unconstrained GANs to controllable GANs} through varying the latent code.

\vspace{2pt}\noindent\textbf{GAN Inversion.}
The generator in GANs typically takes a latent code as the input and outputs a synthesized image.
Hence, it leaves no space for the inference on real images.
To solve this problem, a common practice is to get the reverse mapping from the image space to the latent space, which is also known as GAN Inversion \cite{perarnau2016invertible,zhu2016generative,lipton2017precise,creswell2018inverting,bau2019semantic}.
Prior work either performed instance-level optimization \cite{invertibility,image2stylegan,image2stylegan++} or explicitly learned an encoder to reverse the generator \cite{ali,bigan,lia}.
Some methods combined these two ideas by using the encoder to produce a good starting point for optimization \cite{bau2019seeing,bau2019inverting,zhu2020indomain}.
There are also some works changing the optimization objects by using multiple codes to reconstruct a single image \cite{gu2020image} or optimizing the model weight together with the latent code \cite{pan2020exploiting}.
Being orthogonal to those approaches, our work interprets the latent representation and utilizes GAN inversion as a tool to reuse GAN knowledge for real image editing.

\vspace{2pt}\noindent\textbf{Image-to-Image Translation.}
Image-to-Image translation learns a deterministic model to transfer images from one domain to another.
It is widely used for real image editing considering its fast inference speed \cite{karacan2016learning,park2019semantic,pix2pix,pix2pixhd,liu2017unsupervised,cyclegan}.
Some attempts increased the diversity of the translated images by introducing stochasticity \cite{huang2018multimodal,bicyclegan} or translating images among multiple domains \cite{stargan,stargan2}.
However, all these models rely on paired data or domain labels, which are not easy to obtain.
In this work, we manage to leverage the semantics learned by GANs to create unlimited synthetic data pairs.
We can then apply the knowledge encoded in the latent representation to feed-forward real image editing by training image-to-image translation networks with such synthetic data.

\section{Framework of InterFaceGAN}\label{sec:interfacegan}
In this section, we first provide a rigorous analysis of the properties of the semantics emerging in the latent representation learned by GANs and then construct a pipeline of utilizing the identified semantics for face editing.

\subsection{Semantics in Latent Space}\label{subsec:semantics-interpretation}
Given a well-trained GAN model, the generator can be viewed as a deterministic function $g: \Z\rightarrow\X$.
Here, $\Z\subseteq\R^{d}$ denotes the $d$-dimensional latent space, for which Gaussian distribution $\Norm(\0, \I_d)$ is commonly used \cite{sngan,pggan,biggan,stylegan}.
$\X$ stands for the image space, where each sample $\x$ possesses certain semantic information, like gender and age for face model.
Suppose we have a semantic scoring function $f_S: \X\rightarrow\S$, where $\S\subseteq\R^m$ represents the semantic space with $m$ semantics.
We can bridge the latent space $\Z$ and the semantic space $\S$ with $\s = f_S(g(\z))$, where $\s$ and $\z$ denote semantic scores and the sampled latent code respectively.

\vspace{2pt}\noindent\textbf{Single Semantic.}
It has been widely observed that when linearly interpolating two latent codes, $\z_1$ and $\z_2$, the appearance of the synthesis changes continuously \cite{dcgan,biggan,stylegan}.
It implicitly means that the semantics contained in the image also change gradually.
According to \emph{\textbf{Property~1}}, the interpolation from $\z_1$ to $\z_2$ defines a direction in $\Z$, which further defines a hyperplane.
We therefore make an assumption\footnote{This assumption is empirically demonstrated in Section~\ref{sec:interpretation}.} that for any binary semantic (\emph{e.g.}, male \emph{v.s.} female), there exists a hyperplane in the latent space serving as the separation boundary.
Semantic remains the same when the latent code walks within one side of the hyperplane yet turns into the opposite when across the boundary.

Given a hyperplane with unit normal vector $\n\in\R^d$, we define the ``distance'' from a sample $\z$ to this hyperplane as
\begin{align}
  \dis(\n, \z)= \n^T\z.  \label{eq:distance}
\end{align}
Here, $\dis(\cdot,\cdot)$ is not a strictly defined distance since it can be negative.
When $\z$ lies near the boundary and is moved toward and across the hyperplane, both the ``distance'' and the semantic score vary accordingly.
Moreover, it is just when the ``distance'' changes its numerical sign that the semantic attribute reverses.
We therefore expect these two items to be linearly dependent with
\begin{align}
  f(g(\z)) = \lambda\dis(\n,\z),  \label{eq:linear-dependency}
\end{align}
where $f(\cdot)$ is the scoring function for a particular semantic, and $\lambda > 0$ is a scalar to measure how fast the semantic varies along with the change of ``distance''.
According to \emph{\textbf{Property~2}}, random samples drawn from $\Norm(\0,\I_d)$ are very likely to locate close enough to a given hyperplane.
Therefore, the corresponding semantic can be modeled by the linear subspace that is defined by $\n$.

\vspace{2pt}\emph{\textbf{Property~1}
Given $\n\in\R^d$ with $\n\neq\0$, the set $\{\z\in\R^d:\n^T\z=0\}$ defines a hyperplane in $\R^d$, and $\n$ is called the normal vector. All vectors $\z\in\R^d$ satisfying $\n^T\z>0$ locate from the same side of the hyperplane.
}

\vspace{2pt}\emph{\textbf{Property~2}
Given $\n\in\R^d$ with $\n^T\n=1$, which defines a hyperplane, and a multivariate random variable $\z\sim\Norm(\0,\I_d)$, we have $\Prob(|\n^T\z|\leq2\alpha~\sqrt{\frac{d}{d-2}})\geq(1-3e^{-c d})(1-\frac{2}{\alpha}e^{-\alpha^2/2})$ for any $\alpha\geq1$ and $d\geq4$. Here, $\Prob(\cdot)$ stands for probability and $c$ is a fixed positive constant.\footnote{When $d=512$, we have $P(|\n^T\z|>5.0)<1e^{-6}$. It suggests that almost all sampled latent codes are expected to locate within 5 unit-length to the boundary. Proof can be found in \textbf{Appendix}.}
}

\vspace{5pt}\noindent\textbf{Multiple Semantics.}
When the case comes to $m$ different semantics, we have
\begin{align}
  \s \equiv f_S(g(\z))=\Lambda\N^T\z,  \label{eq:multiple-semantics}
\end{align}
where $\s = [s_1, \dots, s_m]^T$ denotes the semantic scores, $\Lambda = \text{diag}(\lambda_1, \dots, \lambda_m)$ is a diagonal matrix containing the linear coefficients, and $\N = [\n_1, \dots, \n_m]$ indicates the separation boundaries.
Aware of the distribution of random sample $\z$, which is $\Norm(\0,\I_d)$, we can compute the mean and covariance matrix of the semantic scores $\s$ as
\begin{align}
  \bm{\mu}_\s &= \E(\Lambda\N^T\z) = \Lambda\N^T\E(\z) = \0,  \label{eq:score-mean}  \\
  \bm{\Sigma}_\s &= \E(\Lambda\N^T\z\z^T\N\Lambda^T) = \Lambda\N^T\E(\z\z^T)\N\Lambda^T  \notag \\
                 &= \Lambda\N^T\N\Lambda.  \label{eq:score-cov}
\end{align}

We therefore have $\s\sim\Norm(\0,\bm{\Sigma}_\s)$, which is a multivariate normal distribution.
Different entries of $\s$ are disentangled if and only if $\bm{\Sigma}_\s$ is a diagonal matrix, which requires $\{\n_1, \dots, \n_m\}$ to be orthogonal with each other.
If this condition does not hold, some semantics will entangle with each other.
$\n_i^T\n_j$ can be used to measure the entanglement between the $i$-th and $j$-th semantics to some extent.

\subsection{Manipulation in Latent Space}\label{subsec:semantics-manipulation}
In this part, we introduce how to use the semantics found in the latent space for image editing.

\vspace{2pt}\noindent\textbf{Single Attribute Manipulation.}
According to Eq. \eqref{eq:linear-dependency}, to manipulate the attribute of a synthesized image, we can easily edit the original latent code $\z$ with $\z_{edit}=\z + \alpha \n$.
It will make the synthesis look more positive on such semantic with $\alpha>0$ since the score becomes $f(g(\z_{edit}))=f(g(\z)) + \lambda\alpha$ after editing.
Similarly, $\alpha < 0$ will make the synthesis look more negative.

\begin{figure}[t]
  \centering
  \includegraphics[width=0.75\linewidth]{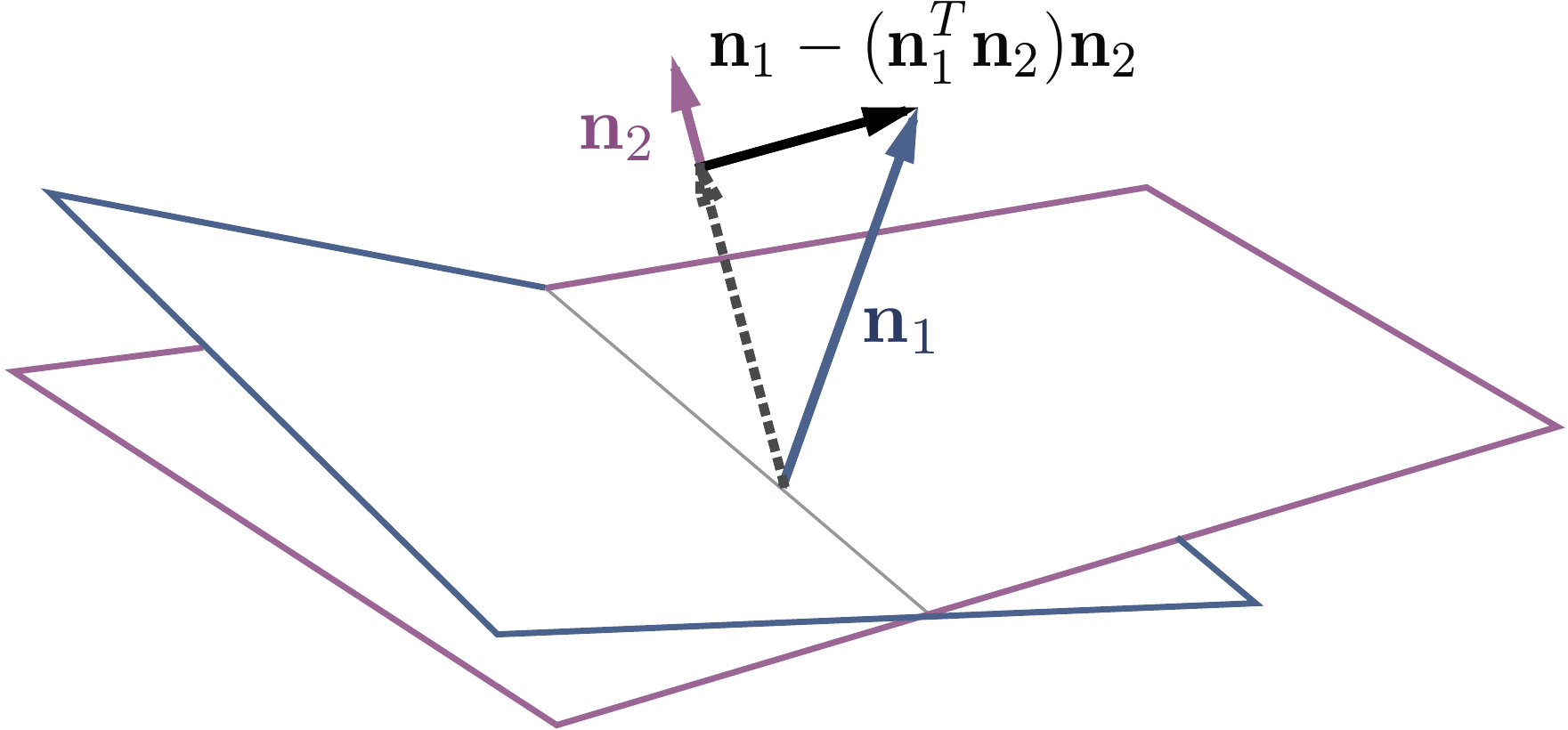}
  \vspace{-8pt}
  \caption{
    Illustration of the \textbf{conditional manipulation via subspace projection}.
    The projection of $\n_1$ onto $\n_2$ is subtracted from $\n_1$, resulting in a new direction $\n_1 - (\n_1^T\n_2)\n_2$.
  }
  \label{fig:subspace}
  \vspace{0pt}
\end{figure}

\vspace{2pt}\noindent\textbf{Conditional Manipulation.}
When there is more than one attribute, editing one may affect another since some semantics can be entangled.
To achieve more precise control, we propose \emph{conditional manipulation} by manually forcing $\N^T\N$ in Eq. \eqref{eq:score-cov} to be diagonal.
In particular, we use projection to make different vectors orthogonal.
As shown in Figure~\ref{fig:subspace}, given two hyperplanes with normal vectors $\n_1$ and $\n_2$, we find a projected direction $\n_1 - (\n_1^T\n_2)\n_2$ such that moving samples along this new direction can change ``attribute 1'' without affecting ``attribute 2''.
If there are multiple attributes to be conditioned on, we subtract the projection from the primal direction onto the plane constructed by all conditioned directions.
Note that the proposed conditional manipulation requires each semantic subspace to be independent of others.
In other words, if two semantics share the same subspace, it is hard to isolate one from the other via subspace projection, but it rarely happens in a high-dimensional space.

\vspace{2pt}\noindent\textbf{Real Image Manipulation.}
InterFaceGAN enables semantic editing from the latent space of a \emph{fixed} GAN model.
To manipulate real images, we should infer the best latent code that can reconstruct the target image, \emph{i.e.}, GAN inversion.
For this purpose, both optimization-based \cite{zhu2020indomain} and learning-based \cite{lia} approaches can be used.
We thoroughly evaluate their strengths and weaknesses in Section~\ref{sec:real-image-manipulation}.
We also use InterFaceGAN to prepare synthetic data pairs and then train image-to-image translation models \cite{pix2pixhd} to achieve real face editing, with the results shown in Section~\ref{sec:real-image-manipulation}.

\subsection{Implementation Details}\label{subsec:implementation-details}
We choose five key facial attributes for analysis, including pose, smile (expression), age, gender, and eyeglasses.
The corresponding positive directions are defined as turning right, laughing, getting old, changing to male, and wearing eyeglasses.
Note that we can always easily plug in more attributes as long as the attribute classifier is available.

To better predict these attributes from synthesized images, we train an auxiliary attribute prediction model using the annotations from the CelebA dataset \cite{celeba} with ResNet-50 network \cite{resnet}.
This model is trained with multi-task losses to simultaneously predict smile, age, gender, eyeglasses, as well as the 5-point facial landmarks (\emph{i.e.}, left eye, right eye, nose, left corner of mouth, right corner of mouth).
Here, the facial landmarks are used to compute yaw pose, which is also treated as a binary attribute (\emph{i.e.}, left \emph{v.s.} right).
Besides the landmarks, all other attributes are learned as a bi-classification problem with softmax cross-entropy loss, while landmarks are optimized with $l_2$ regression loss.
As images produced by PGGAN and StyleGAN are with $1024\times1024$ resolution, we resize them to $224\times224$ before feeding them to the attribute model.

Given the pre-trained GAN model, we synthesize $500K$ images by randomly sampling from the latent space.
There are two main reasons for preparing such a large-scale dataset:
(i) to eliminate the randomness caused by sampling and make sure the distribution of the sampled code is as expected,
and (ii) to get enough wearing-glasses samples, which are rare in CelebA-HQ dataset \cite{pggan}.

To find the semantic boundaries in the latent space, we use the pre-trained attribute prediction model to assign attribute scores for all $500K$ synthesized images.
We sort the corresponding scores for each attribute and choose $10K$ samples with the highest scores and $10K$ with the lowest ones as candidates.
The reason in doing so is that the prediction model is not perfect and may produce wrong predictions for ambiguous samples, \emph{e.g.}, middle-aged person for age attribute.
We then randomly choose 70\% samples from the candidates as the training set to learn a linear SVM, resulting in a decision boundary.
Recall that normal directions of all boundaries are normalized to unit vectors.
The remaining 30\% samples are used for verifying how the linear classifier behaves.
Here, for SVM training, the inputs are the $512d$ latent codes and the predicted attribute scores are treated as binary labels.
Here, note that learning a ``bias'' term in the SVM classifier or not barely affects the learned direction.
That is because we only choose the samples with the highest confidence level for training and they can be easily separated.
We have tried to add the ``bias'' term and the learned bias turns out to be small.

\begin{figure*}[t]
  \centering
  \includegraphics[width=1.0\linewidth]{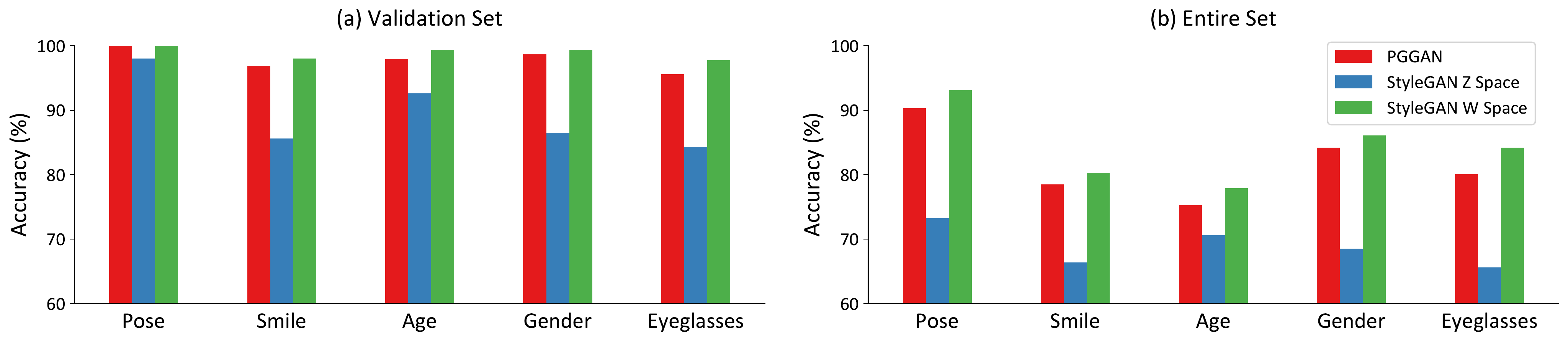}
  \vspace{-22pt}
  \caption{
    \textbf{Classification accuracy} (\%) on the latent separation boundaries of PGGAN \cite{pggan} and StyleGAN \cite{stylegan} with respect to different attributes.
  }
  \label{fig:separation-accuracy}
  \vspace{0pt}
\end{figure*}

\section{Interpreting Face Representation}\label{sec:interpretation}
In this section, we apply InterFaceGAN to interpret the face representation learned by state-of-the-art GAN models, \emph{i.e.}, PGGAN \cite{pggan} and StyleGAN \cite{stylegan}, both of which can produce high-quality faces with $1024\times1024$ resolution.
PGGAN is a representative of the traditional generator where the latent code is only fed into the very first convolutional layer.
By contrast, StyleGAN proposed a style-based generator, which first maps the latent code from latent space $\Z$ to a disentangled latent space $\W$ before applying it for the generation.
In addition, the disentangled latent code is fed to all convolutional layers.\footnote{When feeding $\w$ to each convolutional layer, StyleGAN supports truncation to ensure the synthesis quality. We use truncation 0.7 (1 means no truncation) for the first eight layers in all experiments.}

\subsection{Separability of Latent Space}\label{subsec:latent-space-separation}
Section~\ref{subsec:semantics-interpretation} introduces our assumption that for any binary attribute, there exists a hyperplane in the latent space such that all samples from the same side are with the same attribute.
In this part, we first evaluate this assumption.

\subsubsection{PGGAN}

For each attribute, we will get a latent boundary after the training of the linear SVM classifier.
We evaluate the classification performance on the validation set ($3K$ positive testing samples and $3K$ negative testing samples) and the entire set (remaining $480K$ samples besides the $20K$ candidates with high confidence level).
Figure~\ref{fig:separation-accuracy} shows the results.
We find that all linear boundaries of PGGAN achieve over 95\% accuracy on the validation set and over 75\% on the entire set, suggesting that for a binary attribute, there indeed exists a linear hyperplane in the latent space that can well separate the data into two groups.

\begin{figure}[t]
  \centering
  \includegraphics[width=1.0\linewidth]{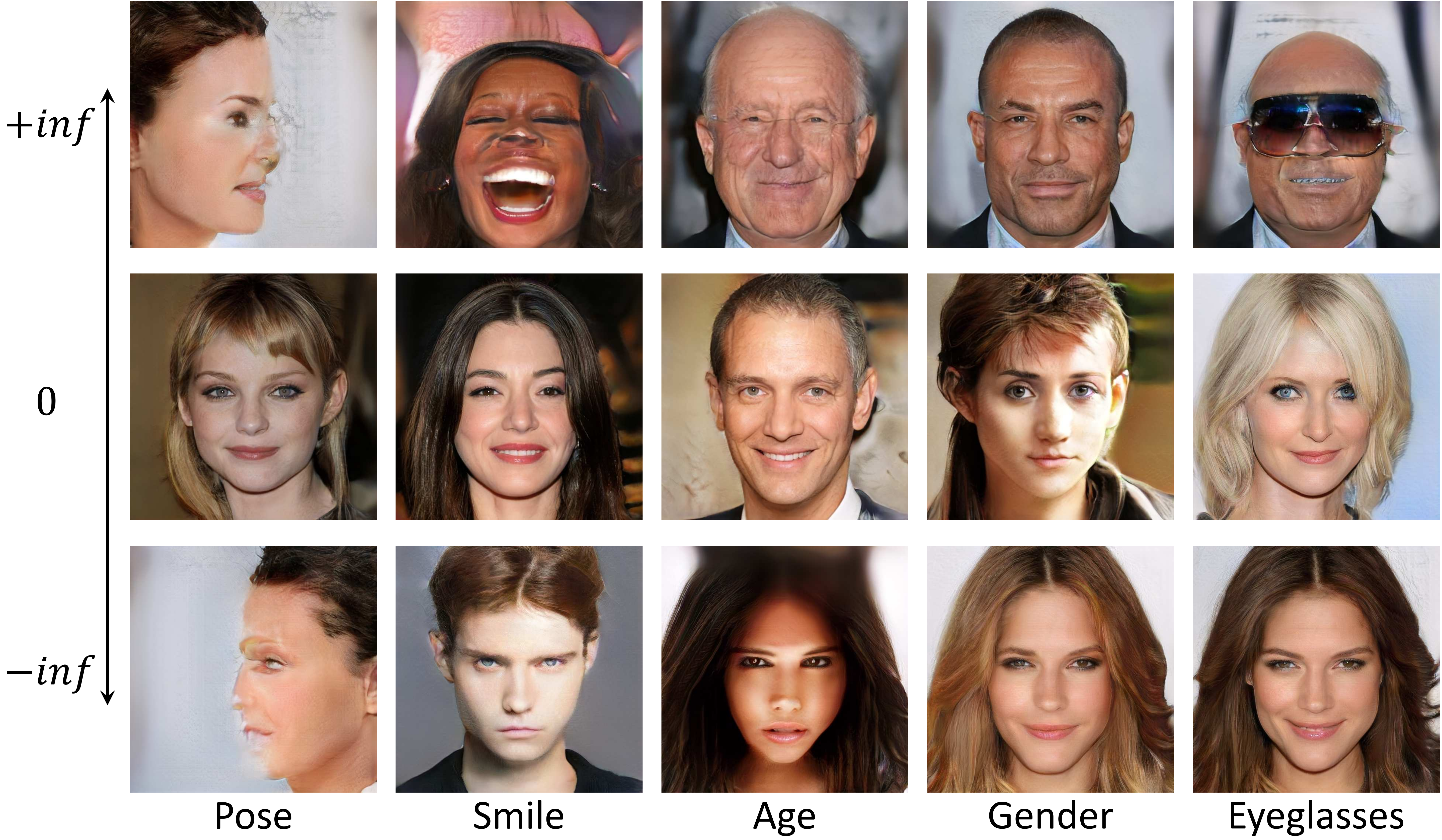}
  \vspace{-20pt}
  \caption{
    Synthesized samples by PGGAN \cite{pggan} with the distance near to (middle row) and extremely far away from (top and bottom rows) the separation boundary.
    Each column corresponds to a particular attribute.
  }
  \label{fig:pggan-separation}
  \vspace{0pt}
\end{figure}

We also visualize some samples in Figure~\ref{fig:pggan-separation} by ranking them by the ``distance'' to the decision boundary.
Those extreme cases (top and bottom rows) are very unlikely to be directly sampled, instead constructed by moving a latent code towards the normal direction ``infinitely''.
Here, to achieve ``zero'' manipulation step, we randomly sample latent codes within the separation hyperplane, while to achieve ``infinite'' manipulation step, we directly use the normal vector $\n$ (or $-\n$) as the sampled code.
From Figure~\ref{fig:pggan-separation}, we can tell that the positive samples and negative samples are distinguishable to each other with respect to the corresponding attribute.
This further demonstrates that the latent space is linearly separable and InterFaceGAN can find the separation hyperplane successfully.

\subsubsection{StyleGAN}

Compared to PGGAN, StyleGAN employs two latent spaces, which are the native latent space $\Z$ and the mapped latent space $\W$.
We analyze the separability of both spaces, and the results are shown in Figure~\ref{fig:separation-accuracy}.
Note that $\W$ space is not subject to normal distribution like $\Z$ space, but we can conduct a similar analysis on $\W$ space, demonstrating the generalization ability of InterFaceGAN.

Figure~\ref{fig:separation-accuracy} shows three observations:
(i) Boundaries from both $\Z$ space and $\W$ space separate the validation set well. Even though the performances of $\Z$ boundaries on the entire set are not satisfying, it may be caused by the inaccurate attribute prediction on the ambiguous samples.
(ii) $\W$ space shows much stronger separability than $\Z$ space. That is because $\w\in\W$, instead of $\z\in\Z$, is the code finally fed into the generator. Accordingly, the generator tends to learn various semantics based on $\W$ space.
(iii) The accuracy of StyleGAN $\W$ space is higher than the PGGAN $\Z$ space. The reason is that the semantic attributes may not be normally distributed in real data. Compared to $\Z$ space, which is subject to the normal distribution, $\W$ space has no constraints and hence is able to better fit the underlying real distribution. This is consistent with the design of $\W$ space in StyleGAN \cite{stylegan}.

\begin{figure*}[t]
  \centering
  \includegraphics[width=1.0\linewidth]{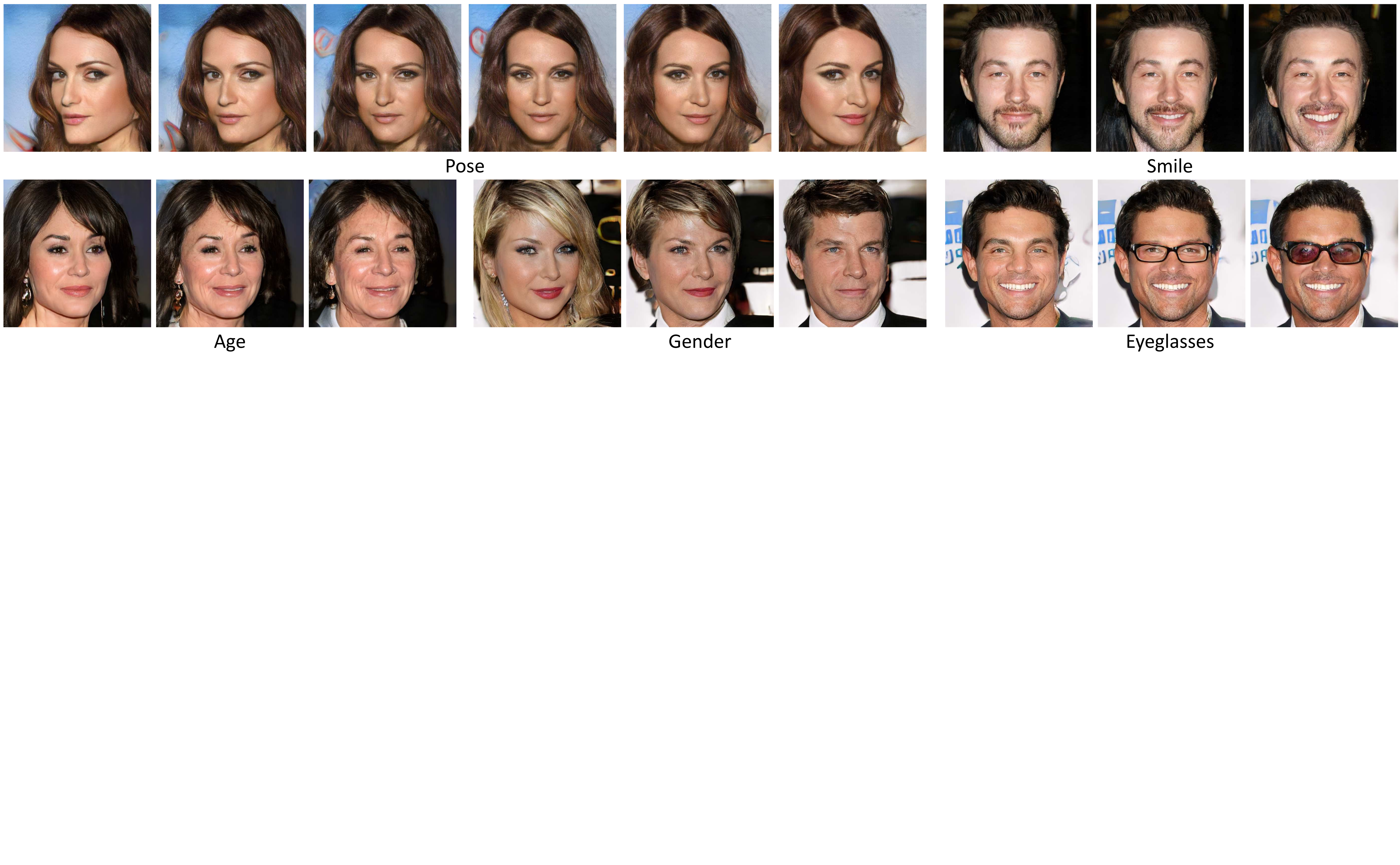}
  \vspace{-22pt}
  \caption{
    \textbf{Single attribute manipulation} results with PGGAN \cite{pggan}.
    The top left shows the same person under gradually changed poses.
    The remaining samples correspond to the results of manipulating four different attributes.
    The central one is the original synthesis for each triplet, while the left and right stand for the results by moving the latent code towards the negative and positive directions respectively.
  }
  \label{fig:pggan-manipulation}
  \vspace{-5pt}
\end{figure*}

\begin{figure*}[t]
  \centering
  \includegraphics[width=1.0\linewidth]{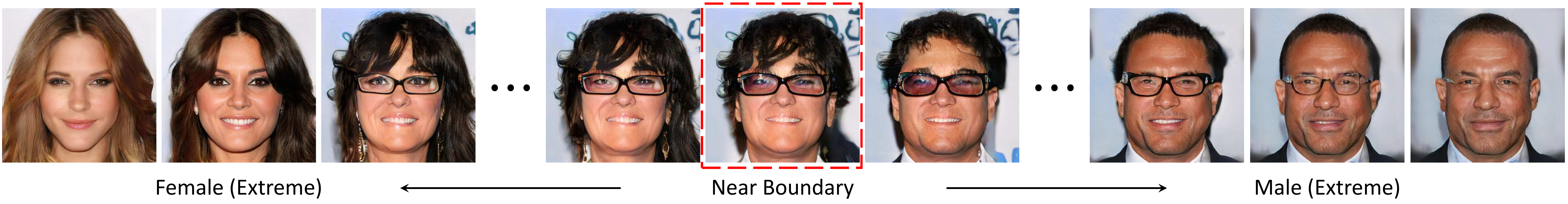}
  \vspace{-22pt}
  \caption{
    Illustration of the \textbf{distance effect} by taking gender manipulation with PGGAN \cite{pggan} as an example.
    The image in the red dashed box stands for the original synthesis.
    Our approach performs well when the latent code locates close to the boundary.
    However, when the distance keeps increasing, the synthesized images are no longer like the same person.
  }
  \label{fig:pggan-limitation}
  \vspace{-5pt}
\end{figure*}

\subsection{Semantics in Latent Space for Face Manipulation}\label{subsec:latent-space-manipulation}
In this part, we verify whether the semantics found by InterFaceGAN are manipulable.

\subsubsection{PGGAN}

\vspace{2pt}\noindent\textbf{Manipulating Single Attribute.}
Figure~\ref{fig:pggan-manipulation} plots the manipulation results on five different attributes.
It suggests that our manipulation approach performs well on all attributes in both positive and negative directions.
Particularly on ``pose'' attribute, we observe that even the boundary is searched by solving a bi-classification problem, moving the latent code can produce continuous change.
Furthermore, although there lacks enough data with extreme poses in the training set, GAN can hallucinate how profile faces should look.
The same situation also appears on the eyeglasses attribute.
We can manually create many faces wearing eyeglasses despite the inadequate data in the training set.
These two observations provide strong evidence that GAN does not produce images randomly, but learns some interpretable semantics in the latent space.

\vspace{2pt}\noindent\textbf{Distance Effect of Semantic Subspace.}
When manipulating the latent code, we observe a noticeable distance effect: the samples will suffer from severe changes in appearance if being moved too far from the boundary, and finally tend to become the extreme cases shown in Figure~\ref{fig:pggan-separation}.
Figure~\ref{fig:pggan-limitation} illustrates this phenomenon by taking gender editing as an instance.
Near-boundary manipulation works well.
When samples go beyond a certain region, however, the editing results are no longer like the original face anymore.
However, this effect does not affect our understanding of the disentangled semantics in the latent space.
That is because such extreme samples are unlikely to be directly drawn from a standard normal distribution, which is pointed out in \emph{\textbf{Property~2}} in Section~\ref{subsec:semantics-interpretation}.
Instead, they are constructed manually by keeping moving a normally sampled latent code along a certain direction.

\begin{figure}[t]
  \centering
  \includegraphics[width=1.0\linewidth]{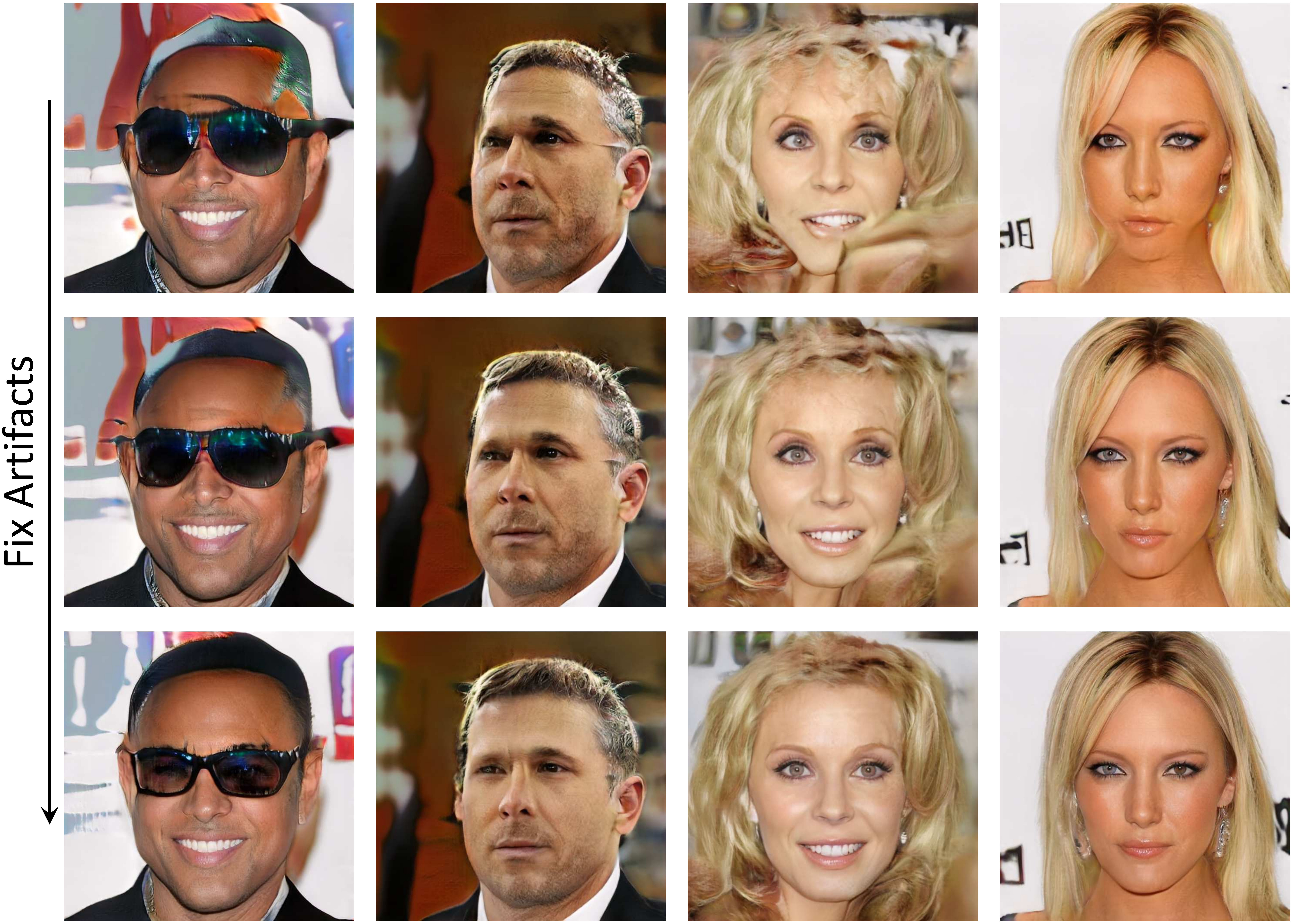}
  \vspace{-20pt}
  \caption{
    Examples on \textbf{fixing the artifacts} that PGGAN \cite{pggan} made.
    The first row shows several bad synthesis results, while the following two rows present the gradually corrected synthesis by moving the latent codes towards the positive ``quality'' direction.
  }
  \label{fig:pggan-correction}
  \vspace{-5pt}
\end{figure}

\vspace{2pt}\noindent\textbf{Fixing Artifacts.}
We further apply our approach to fix the artifacts that sometimes occur in the synthesis.
We manually labelled $4K$ deficient synthesis and then trained a linear SVM to find the separation hyperplane.
We surprisingly find that GAN also encodes such quality information in the latent space.
Based on this discovery, we manage to correct some mistakes GAN has made in the generation process by moving the latent code towards the positive ``quality'' direction, as shown in Figure~\ref{fig:pggan-correction}.

\begin{figure*}[t]
  \centering
  \includegraphics[width=1.0\linewidth]{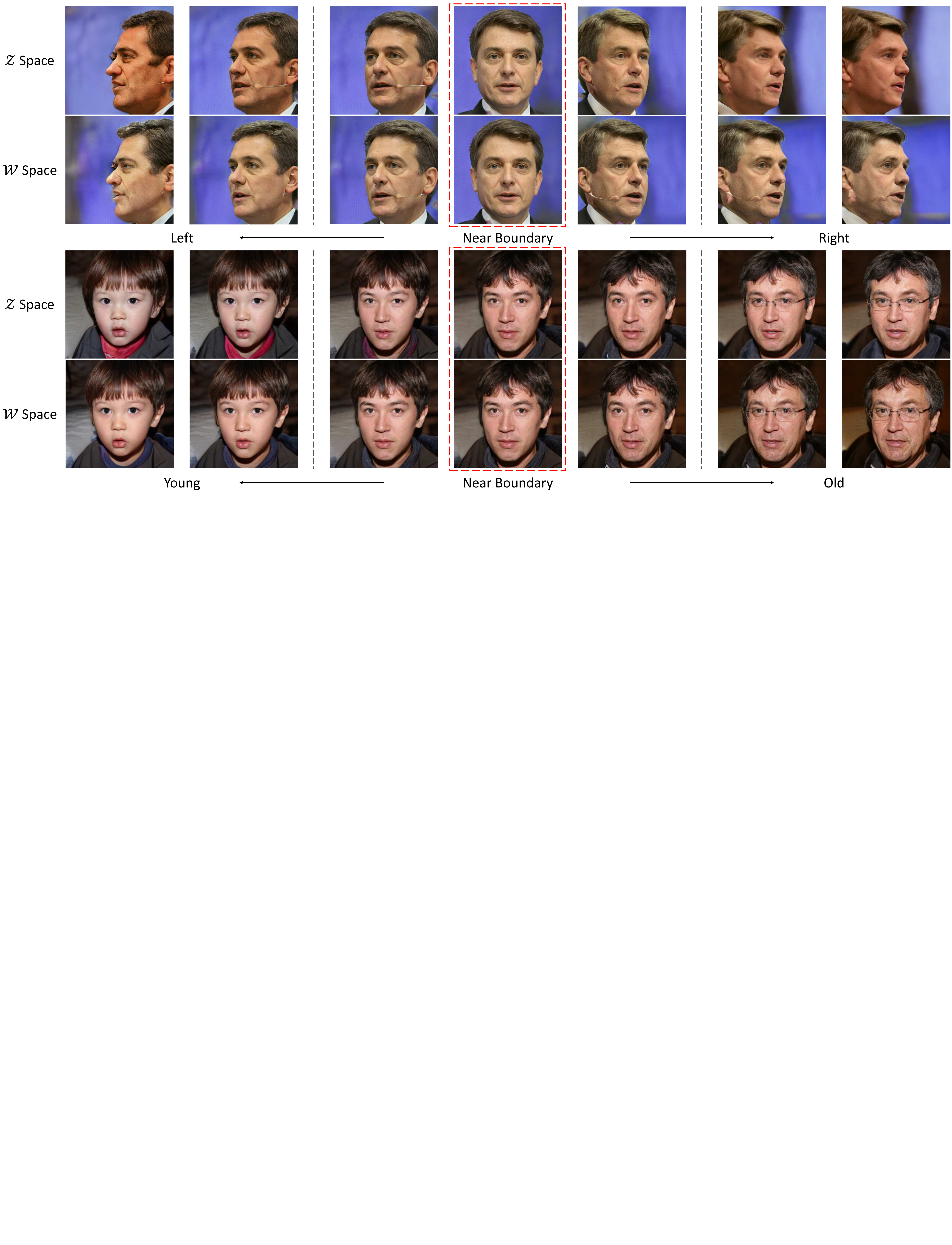}
  \vspace{-20pt}
  \caption{
    \textbf{Attribute editing} results on StyleGAN \cite{stylegan}.
    For two attributes pose and age, the top row shows the manipulation results with respect to $\Z$ space, whilst the bottom row corresponds to $\W$ space.
    Images in red dashed boxes represent the original synthesis.
    Images between two black dashed lines stand for near-boundary manipulation, and the other images stand for long-distance manipulation.
  }
  \label{fig:stylegan-manipulation}
  \vspace{-5pt}
\end{figure*}

\subsubsection{StyleGAN}

We further apply InterFaceGAN to StyleGAN by manipulating the latent codes in both $\Z$ space and $\W$ space.
Figure~\ref{fig:stylegan-manipulation} shows the following observations:
(i) InterFaceGAN works well on the style-based generator. We manage to edit the attributes by altering the latent code in either $\Z$ space or $\W$ space.
(ii) By learning from a more diverse dataset, FF-HQ~\cite{stylegan}, StyleGAN learns various semantics more thoroughly. For example, StyleGAN can produce high-quality profile faces (first example) and generate children when making people younger (second example). This is beyond the ability of PGGAN, which is trained on CelebA-HQ dataset \cite{pggan}.
(iii) Some attributes correlate with each other. For example, people are wearing eyeglasses when turning old (second example). A more detailed analysis of this phenomenon will be discussed in Section~\ref{subsec:conditional-manipulation}.
(iv) $\W$ space shows better performance than $\Z$ space, especially for long-distance manipulation. In other words, when the latent code locates near the separation boundary, manipulations in $\Z$ space and $\W$ space have a similar effect. However, when the latent code goes further from the boundary, manipulating one attribute in $\Z$ space might affect another. Taking pose editing (first example) as an instance, the skin color of the target person varies when moving the latent code along the pose direction. By contrast, $\W$ space shows stronger robustness.

\subsection{Disentanglement Analysis and Conditional Manipulation}\label{subsec:conditional-manipulation}
This section discusses the disentanglement between different semantics encoded in the latent representation and evaluates the conditional manipulation approach.

\setlength{\tabcolsep}{1pt}
\begin{table*}[t]
  \captionsetup[subfloat]{captionskip=2pt,position=top}
  \caption{
    \textbf{Disentanglement analysis} on PGGAN \cite{pggan}.
  }
  \label{tab:pggan-disentanglement}
  \vspace{-15pt}
  \centering\footnotesize
  \subfloat[Training data.]{
    \begin{tabular}{c*{\NumAttributes}{C{24pt}}}
      \multicolumn{1}{c}{} &
      \multicolumn{1}{c}{\rot{Pose}} &
      \multicolumn{1}{c}{\rot{Smile}} &
      \multicolumn{1}{c}{\rot{Age}} &
      \multicolumn{1}{c}{\rot{Gender}} &
      \multicolumn{1}{c}{\rot{Glasses}}             \\
      Pose    & 1.00 & -0.01 & 0.00 &  0.00 &  0.01 \\
      Smile   &      &  1.00 & 0.03 & -0.09 & -0.03 \\
      Age     &      &       & 1.00 &  0.20 &  0.15 \\
      Gender  &      &       &      &  1.00 &  0.19 \\
      Glasses &      &       &      &       &  1.00 \\
    \end{tabular}
  }
  \hspace{5pt}
  \subfloat[Synthesized data.]{
    \begin{tabular}{c*{\NumAttributes}{C{24pt}}}
      \multicolumn{1}{c}{} &
      \multicolumn{1}{c}{\rot{Pose}} &
      \multicolumn{1}{c}{\rot{Smile}} &
      \multicolumn{1}{c}{\rot{Age}} &
      \multicolumn{1}{c}{\rot{Gender}} &
      \multicolumn{1}{c}{\rot{Glasses}}              \\
      Pose    & 1.00 & -0.01 & -0.01 & -0.02 &  0.00 \\
      Smile   &      &  1.00 &  0.02 & -0.08 & -0.01 \\
      Age     &      &       &  1.00 &  0.42 &  0.35 \\
      Gender  &      &       &       &  1.00 &  0.47 \\
      Glasses &      &       &       &       &  1.00 \\
    \end{tabular}
  }
  \hspace{5pt}
  \subfloat[Semantic boundaries.]{
    \begin{tabular}{c*{\NumAttributes}{C{24pt}}}
      \multicolumn{1}{c}{} &
      \multicolumn{1}{c}{\rot{Pose}} &
      \multicolumn{1}{c}{\rot{Smile}} &
      \multicolumn{1}{c}{\rot{Age}} &
      \multicolumn{1}{c}{\rot{Gender}} &
      \multicolumn{1}{c}{\rot{Glasses}}              \\
      Pose    & 1.00 & -0.04 & -0.06 & -0.05 & -0.04 \\
      Smile   &      &  1.00 &  0.04 & -0.10 & -0.05 \\
      Age     &      &       &  1.00 &  0.49 &  0.38 \\
      Gender  &      &       &       &  1.00 &  0.52 \\
      Glasses &      &       &       &       &  1.00 \\
    \end{tabular}
  }
  \vspace{-5pt}
\end{table*}

\begin{figure*}[t]
  \centering
  \includegraphics[width=1.0\linewidth]{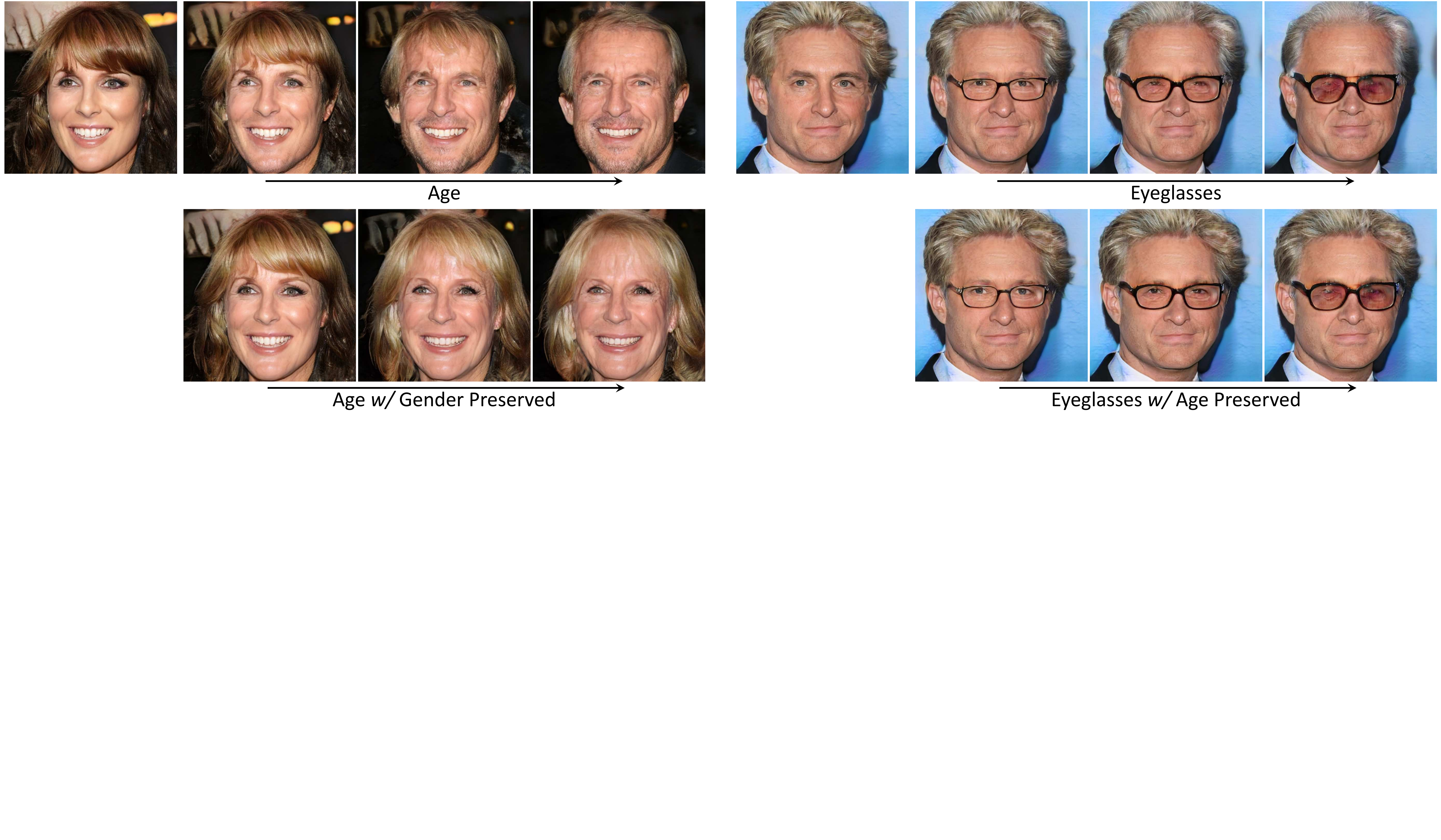}
  \vspace{-20pt}
  \caption{
    \textbf{Conditional manipulation} results using PGGAN \cite{pggan}.
    Left: Manipulating age attribute by preserving gender.
    Right: Manipulating eyeglasses attribute by preserving age.
    For each example, the top row shows the unconditional editing results while the bottom row shows the conditional manipulation.
  }
  \label{fig:pggan-condition}
  \vspace{-5pt}
\end{figure*}

\subsubsection{Disentanglement Measurement}

Besides how one semantic can be well encoded in the latent space, we also study the correlation between multiple semantics and examine the way to decouple correlated attributes.
In particular, we use the following metrics for analysis.
\begin{enumerate}
  \item \emph{Attribute correlation of real data}. We use the prepared predictors to predict the attribute scores of real data on which the GAN model is trained. Then we compute the correlation coefficient  $\rho$ between any two attributes with $\rho_{A_1,A_2}=\frac{Cov(A_1,A_2)}{\sigma_{A_1}\sigma_{A_2}}$. Here, $A_1$ and $A_2$ represent two random variables with respect to these two attributes. $Cov(\cdot,\cdot)$ and $\sigma$ denote covariance and standard deviation respectively.
  \item \emph{Attribute correlation of synthesized data}. We also compute the attribute correlation score on the $500K$ synthesized data. By comparing this score to that of the real data, we can get clues on how GANs learn to encode such semantic knowledge in the latent representation.
  \item \emph{Semantic boundary correlation}. Given any two semantics, with the latent boundaries $\n_1$ and $\n_2$, we compute the cosine similarity between these two directions with $\cos(\n_1, \n_2)=\n_1^T \n_2$. Here, $\n_1$ and $\n_2$ are both unit vectors. This metric is used to evaluate how the above attribute correlation is reflected in our InterFaceGAN framework.
\end{enumerate}

\subsubsection{PGGAN}

\vspace{2pt}\noindent\textbf{Disentanglement Analysis.}
Table~\ref{tab:pggan-disentanglement} reports the correlation metrics of PGGAN model trained on CelebA-HQ dataset ~\cite{pggan}.
By comparing the attribution correlation of real data (Table~\ref{tab:pggan-disentanglement}~(a)) and that of synthesized data (Table~\ref{tab:pggan-disentanglement}~(b)), we can tell that they are very close to each other.
For example, ``pose'' and ``smile'' are almost independent to other attributes, while ``gender'', ``age'', and ``eyeglasses'' are highly correlated with each other from both real data and synthesized data.
For example, elder men are more likely to wear eyeglasses.
This implies that GANs learn the underlying semantic distribution from real observation when trained to synthesize images.
Then, by comparing such attribution correlation with the boundary correlation (Table~\ref{tab:pggan-disentanglement}~(c)), we also find that they behave similarly.
This suggests that InterFaceGAN is able to not only accurately identify the semantics encoded in that latent representation but also capture the entanglement among them.

\begin{figure}[t]
  \centering
  \includegraphics[width=1.0\linewidth]{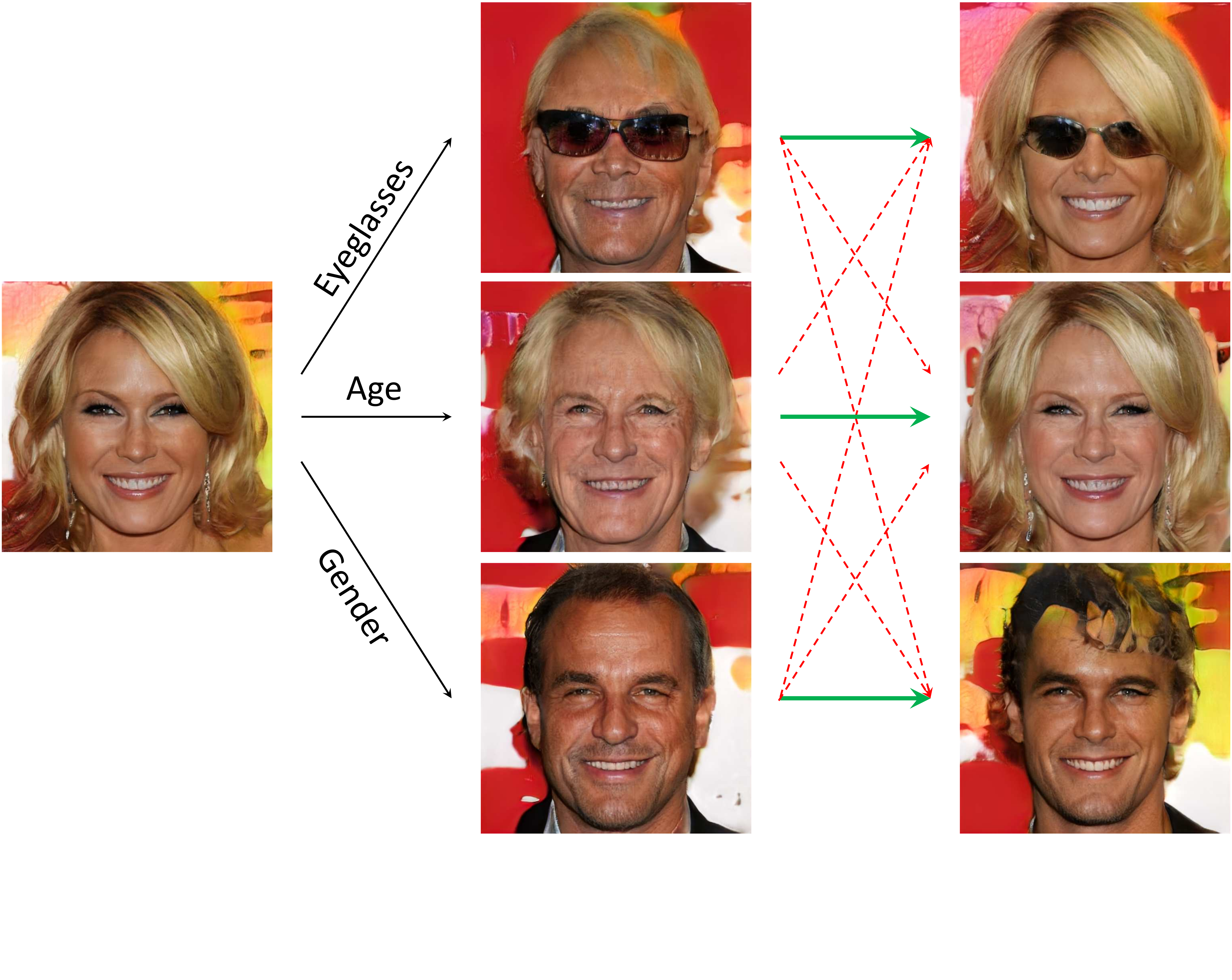}
  \vspace{-20pt}
  \caption{
    \textbf{Conditional manipulation with more than one conditions} using PGGAN \cite{pggan}.
    Left: Original synthesis.
    Middle: Manipulations along a single boundary.
    Right: Conditional manipulations.
    Green arrows indicate the primal direction and red arrows indicate the directions to be conditioned on.
  }
  \label{fig:pggan-multiple-conditions}
  \vspace{-5pt}
\end{figure}

\vspace{2pt}\noindent\textbf{Conditional Manipulation.}
As shown in Table~\ref{tab:pggan-disentanglement}~(c), if two boundaries are not orthogonal to each other, modulating the latent code along one direction will affect the other.
Hence, we propose conditional manipulation via subspace projection to reduce this entanglement as much as possible.
Details are described in Section~\ref{subsec:semantics-manipulation}.
Figure~\ref{fig:pggan-condition} shows the discrepancies between unconditional manipulation and conditional manipulation.
Taking the top-left example in Figure~\ref{fig:pggan-condition} as an instance, the results tend to become male when being edited to get old (top row).
We address this issue by subtracting its projection onto the gender direction from the age direction, resulting in a new direction.
By moving latent codes along this projected direction, we can make sure the gender component is barely affected in the editing process (bottom row).
Figure~\ref{fig:pggan-multiple-conditions} shows a more complicated case where we perform manipulation with multiple conditions.
Taking ``eyeglasses'' attribute as an example, in the beginning, adding eyeglasses is entangled with changing both age and gender.
But we manage to disentangle eyeglasses from age and gender by forcing the eyeglasses direction to be orthogonal to the other two.
Note that the hair region of the bottom right result in Figure~\ref{fig:pggan-multiple-conditions} is mixed with the background.
That is because hair affects ``gender'', but only using ``eyeglasses'' and ``age'' as conditions are hard to demonstrate the demarcation between hair and background.
This may be improved if we can identify another subspace using a background predictor and use it as an additional condition.
Even so, the results in Figure~\ref{fig:pggan-condition} and Figure~\ref{fig:pggan-multiple-conditions} demonstrate that our proposed conditional approach helps to achieve precise attribute control.

\setlength{\tabcolsep}{1pt}
\begin{table*}[t]
  \captionsetup[subfloat]{captionskip=2pt,position=top}
  \caption{
    \textbf{Disentanglement analysis} on StyleGAN \cite{stylegan}.
  }
  \label{tab:stylegan-disentanglement}
  \vspace{-15pt}
  \centering\footnotesize
  \subfloat[Training data.]{
    \begin{tabular}{c*{\NumAttributes}{C{24pt}}}
      \multicolumn{1}{c}{} &
      \multicolumn{1}{c}{\rot{Pose}} &
      \multicolumn{1}{c}{\rot{Smile}} &
      \multicolumn{1}{c}{\rot{Age}} &
      \multicolumn{1}{c}{\rot{Gender}} &
      \multicolumn{1}{c}{\rot{Glasses}}             \\
      Pose    & 1.00 & -0.02 & 0.00 & -0.02 & -0.01 \\
      Smile   &      &  1.00 & 0.01 & -0.17 & -0.03 \\
      Age     &      &       & 1.00 &  0.14 &  0.21 \\
      Gender  &      &       &      &  1.00 &  0.20 \\
      Glasses &      &       &      &       &  1.00 \\
    \end{tabular}
  }
  \hspace{5pt}
  \subfloat[Semantic boundaries from $\Z$ space.]{
    \begin{tabular}{c*{\NumAttributes}{C{24pt}}}
      \multicolumn{1}{c}{} &
      \multicolumn{1}{c}{\rot{Pose}} &
      \multicolumn{1}{c}{\rot{Smile}} &
      \multicolumn{1}{c}{\rot{Age}} &
      \multicolumn{1}{c}{\rot{Gender}} &
      \multicolumn{1}{c}{\rot{Glasses}}              \\
      Pose    & 1.00 & -0.03 &  0.03 & -0.01 & -0.08 \\
      Smile   &      &  1.00 & -0.28 & -0.42 & -0.20 \\
      Age     &      &       &  1.00 &  0.33 &  0.72 \\
      Gender  &      &       &       &  1.00 &  0.44 \\
      Glasses &      &       &       &       &  1.00 \\
    \end{tabular}
  }
  \hspace{5pt}
  \subfloat[Semantic boundaries from $\W$ space.]{
    \begin{tabular}{c*{\NumAttributes}{C{24pt}}}
      \multicolumn{1}{c}{} &
      \multicolumn{1}{c}{\rot{Pose}} &
      \multicolumn{1}{c}{\rot{Smile}} &
      \multicolumn{1}{c}{\rot{Age}} &
      \multicolumn{1}{c}{\rot{Gender}} &
      \multicolumn{1}{c}{\rot{Glasses}}            \\
      Pose    & 1.00 & 0.00 & 0.02 &  0.03 & -0.03 \\
      Smile   &      & 1.00 & 0.03 & -0.06 &  0.02 \\
      Age     &      &      & 1.00 &  0.07 &  0.05 \\
      Gender  &      &      &      &  1.00 &  0.00 \\
      Glasses &      &      &      &       &  1.00 \\
    \end{tabular}
  }
  \vspace{-5pt}
\end{table*}

\begin{figure*}[t]
  \centering
  \includegraphics[width=1.0\linewidth]{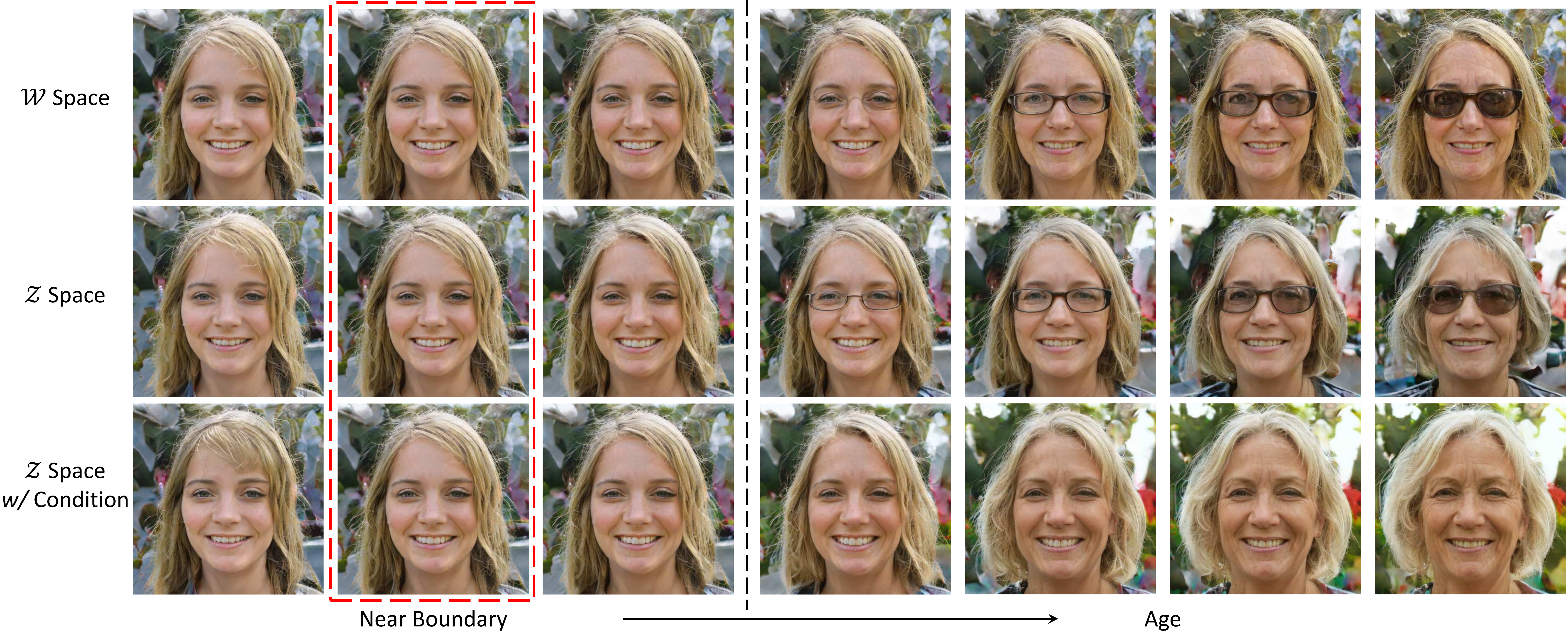}
  \vspace{-20pt}
  \caption{
    \textbf{Conditional manipulation} analysis on StyleGAN \cite{stylegan} by taking age manipulation as an example.
    Images in red dashed boxes represent the original synthesis, while the others show the manipulation process.
    $\W$ space is more disentangled than $\Z$ space, especially for long-distance manipulation, but the entanglement in $\Z$ space can be diminished by the proposed conditional manipulation.
  }
  \label{fig:stylegan-condition}
  \vspace{-5pt}
\end{figure*}

\subsubsection{StyleGAN}

\vspace{2pt}\noindent\textbf{Disentanglement Analysis.}
We conduct a similar analysis on the StyleGAN model trained on FF-HQ dataset \cite{stylegan}.
As mentioned above, StyleGAN introduces a disentangled latent space $\W$ beyond the original latent space $\Z$.
Hence, we analyze the boundary correlation from both of these two spaces.
Results are shown in Table~\ref{tab:stylegan-disentanglement}.
Besides the conclusions from PGGAN, we have three more observations.
(i) ``Smile'' and ``gender'' are not correlated in CelebA-HQ dataset (Table~\ref{tab:pggan-disentanglement}~(a)), but entangled in FF-HQ dataset (Table~\ref{tab:stylegan-disentanglement}~(a)).
Such ``data bias'' may lead to a different latent representation.
(ii) $\W$ space (Table~\ref{tab:stylegan-disentanglement}~(c)) is indeed more disentangled than $\Z$ space (Table~\ref{tab:stylegan-disentanglement}~(b)), which is consistent with the conclusion from Karras \emph{et al.} \cite{stylegan}. In $\W$ space, almost all boundaries are orthogonal to each other.
(iii) The boundary correlation from $\W$ space no longer aligns with the semantic distribution from real data (Table~\ref{tab:stylegan-disentanglement}~(a)). That may be because $\W$ space is subject to a more complicated distribution than Gaussian distribution and the boundaries found by linear classifiers may not be very accurate. Using a non-linear classifier may solve this problem.

\vspace{2pt}\noindent\textbf{Conditional Manipulation.}
We also evaluate the proposed conditional manipulation on the style-based generator to verify its generalization ability.
In particular, given a sample, we manipulate its attribute from both $\Z$ space and $\W$ space, and then perform conditional manipulation in $\Z$ space.
Note that such conditional operation is \emph{not} applicable to $\W$ space.
As shown in Table~\ref{tab:stylegan-disentanglement}~(c), all boundaries are almost orthogonal to each other.
As a result, projection barely changes the primal direction.
Figure~\ref{fig:stylegan-condition} gives an example about the entanglement between ``age'' and ``eyeglasses''.
In Figure~\ref{fig:stylegan-condition}, manipulating from $\Z$ space and $\W$ space produces similar results when the latent code still locates near the boundary.
For long-distance manipulation, $\W$ space (first row) shows superiority over $\Z$ space (second row), \emph{e.g.}, hair length and face shape do not change in the first row.
Even so, ``age'' and ``eyeglasses'' are still entangled in both spaces.
However, we can use subspace projection to decorrelate ``eyeglasses'' from ``age'' in $\Z$ space (third row), resulting in more precise control.
This demonstrates the effectiveness of the proposed conditional manipulation approach.

\setlength{\tabcolsep}{1pt}
\begin{table*}[t]
  \centering\footnotesize
  \captionsetup[subfloat]{captionskip=2pt,position=top}
  \caption{
    \textbf{Re-scoring analysis} on the semantic manipulation achieved by InterFaceGAN.
    Each row shows the results by manipulating a particular attribute.
  }
  \label{tab:rescoring}
  \vspace{-18pt}
  \subfloat[PGGAN \cite{pggan}.]{
    \begin{tabular}{c*{\NumAttributes}{C{24pt}}}
      \multicolumn{1}{c}{} &
      \multicolumn{1}{c}{\rot{Pose}} &
      \multicolumn{1}{c}{\rot{Smile}} &
      \multicolumn{1}{c}{\rot{Age}} &
      \multicolumn{1}{c}{\rot{Gender}} &
      \multicolumn{1}{c}{\rot{Glasses}}               \\
      Pose    &  0.53 &  0.05 & -0.09 & -0.06 & -0.01 \\
      Smile   & -0.01 &  0.60 &  0.03 & -0.07 & -0.01 \\
      Age     & -0.03 & -0.03 &  0.50 &  0.35 &  0.08 \\
      Gender  & -0.02 & -0.07 &  0.20 &  0.59 &  0.08 \\
      Glasses & -0.01 &  0.00 &  0.19 &  0.37 &  0.24 \\
    \end{tabular}
  }
  \hspace{5pt}
  \subfloat[StyleGAN $\Z$ space.]{
    \begin{tabular}{c*{\NumAttributes}{C{24pt}}}
      \multicolumn{1}{c}{} &
      \multicolumn{1}{c}{\rot{Pose}} &
      \multicolumn{1}{c}{\rot{Smile}} &
      \multicolumn{1}{c}{\rot{Age}} &
      \multicolumn{1}{c}{\rot{Gender}} &
      \multicolumn{1}{c}{\rot{Glasses}}               \\
      Pose    &  0.47 &  0.03 & -0.04 & -0.01 & -0.03 \\
      Smile   & -0.01 &  0.50 & -0.17 & -0.21 & -0.07 \\
      Age     &  0.00 & -0.11 &  0.56 &  0.18 &  0.32 \\
      Gender  &  0.01 & -0.24 &  0.27 &  0.45 &  0.13 \\
      Glasses & -0.03 & -0.06 &  0.37 &  0.24 &  0.40 \\
    \end{tabular}
  }
  \hspace{5pt}
  \subfloat[StyleGAN $\W$ space.]{
    \begin{tabular}{c*{\NumAttributes}{C{24pt}}}
      \multicolumn{1}{c}{} &
      \multicolumn{1}{c}{\rot{Pose}} &
      \multicolumn{1}{c}{\rot{Smile}} &
      \multicolumn{1}{c}{\rot{Age}} &
      \multicolumn{1}{c}{\rot{Gender}} &
      \multicolumn{1}{c}{\rot{Glasses}}              \\
      Pose    & 0.51 &  0.02 & -0.06 &  0.01 & -0.03 \\
      Smile   & 0.00 &  0.50 & -0.15 & -0.09 & -0.02 \\
      Age     & 0.01 & -0.02 &  0.54 &  0.03 &  0.20 \\
      Gender  & 0.03 & -0.07 &  0.01 &  0.45 &  0.05 \\
      Glasses & 0.00 &  0.01 &  0.09 &  0.03 &  0.41 \\
    \end{tabular}
  }
  \vspace{-5pt}
\end{table*}

\begin{figure*}[t]
  \centering
  \includegraphics[width=1.0\linewidth]{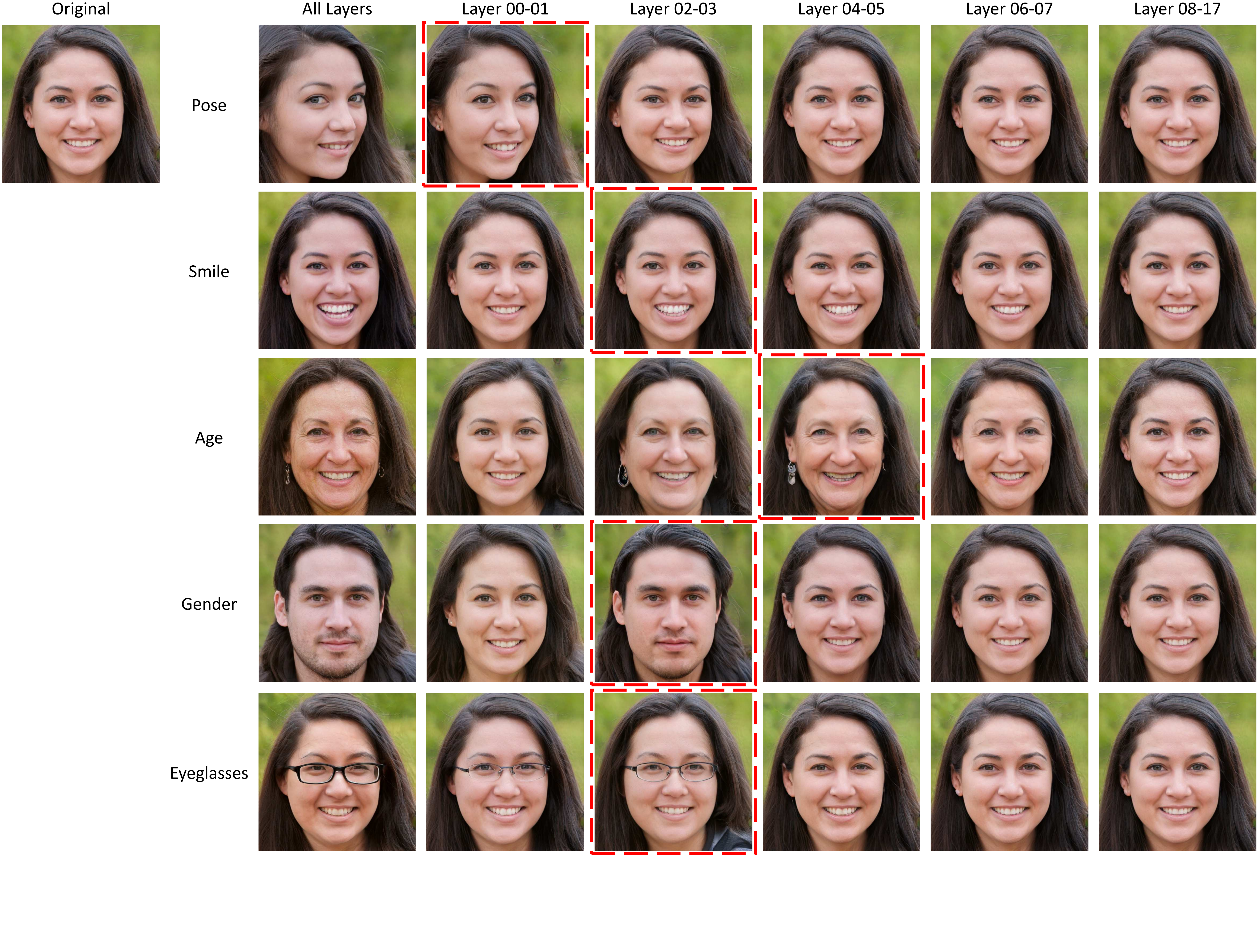}
  \vspace{-22pt}
  \caption{
    \textbf{Layer-wise manipulation} results with StyleGAN \cite{stylegan}.
    On the top-left corner is the raw synthesis. The images on the first column are the results of varying the codes at all the layers.
    Images in red dashed boxes highlight the best layer-wise manipulation results.
  }
  \label{fig:layerwise}
  \vspace{-5pt}
\end{figure*}

\section{Quantitative Analysis on Manipulation}\label{sec:quantitative-analysis}
We have shown the qualitative results in Section~\ref{subsec:latent-space-manipulation} and Section~\ref{subsec:conditional-manipulation} to exhibit the controllable disentangled semantics identified by InterFaceGAN.
In this part, we quantitatively analyze the properties of these disentangled semantics as well as the face manipulation approach.
We conduct the following analysis:
(i) whether the manipulation can indeed increase or decrease the attribute score, and how manipulating one attribute affects the scores of other attributes (Section~\ref{subsec:rescoring-analysis});
(ii) how GANs learn the face representation layer by layer (Section~\ref{subsec:layerwise-analysis});
and (iii) how the attribute manipulation affects the face identity (Section~\ref{subsec:identity-analysis}).

\subsection{Evaluating Editing Performance with Re-scoring}\label{subsec:rescoring-analysis}
After manipulation, we further predict the attribute scores of the resulted face.
In this way, we can compute the score change compared to the input face.
We use the re-scoring to verify whether the manipulation has been successfully performed.
For example, when we move the latent code towards ``male'' direction (\emph{i.e.}, the positive direction of ``gender'' boundary), we would expect the ``gender'' score to increase.
This metric can also be used to evaluate the disentanglement between different semantics.
For example, if we want to see how ``gender'' and ``age'' correlate with each other, we can move the latent code along the ``gender'' boundary and see how the ``age'' score varies.

We use $2K$ synthesis for re-scoring analysis on PGGAN~\cite{pggan}, StyleGAN~\cite{stylegan} $\Z$ space, and StyleGAN $\W$ space. The results in Table~\ref{tab:rescoring} indicate that:
(i) InterFaceGAN can convincingly increase the target semantic scores by manipulating the appropriate attributes (see diagonal entries).
(ii) Manipulating one attribute may affect the scores of other attributes. Taking PGGAN (Table~\ref{tab:rescoring}~(a)) as an example, when manipulating ``age'', ``gender'' score also increases. This is consistent with the observation from Section~\ref{subsec:conditional-manipulation}. To some extent, we can treat this as an another disentanglement measurement. Under this metric, we also see that $\W$ space (Table~\ref{tab:rescoring}~(c)) is more disentangled than $\Z$ space (Table~\ref{tab:rescoring}~(b)) in StyleGAN.
(iii) This new metric is asymmetric. Taking PGGAN (Table~\ref{tab:rescoring}~(a)) as an example, when we manipulate ``age'', ``eyeglasses'' is barely affected. But when we manipulate ``eyeglasses'', ``age'' score increase a lot. The same phenomenon also happens to ``gender'' and ``eyeglasses''. This provides us with adequate information about the entanglement between the semantics learned in the latent representation.

\subsection{Per-Layer Representation Learned by GANs}\label{subsec:layerwise-analysis}
Unlike the traditional generator, StyleGAN \cite{stylegan} feeds the latent code to all convolutional layers.
This enables us to study the per-layer representation.
Given a boundary, we can use it to only vary the latent codes that are fed into some particular layers.
In practice, we divide the 18 layers into 5 groups, \emph{i.e.}, 00-01, 02-03, 04-05, 06-07, and 08-17.
Then we conduct the same experiment as in re-scoring analysis to examine the importance of each group for each attribute.

\setlength{\tabcolsep}{1pt}
\begin{table}[t]
  \setlength{\tabcolsep}{5.5pt}
  \caption{
    \textbf{Layer-wise analysis} on the semantic manipulation achieved by InterFaceGAN using StyleGAN \cite{stylegan}.
    Each row shows the results by manipulating a particular attribute and how the scores of itself and others vary.
  }
  \label{tab:layerwise}
  \vspace{-8pt}
  \centering\small
  \begin{tabular}{c|c*{5}{C{24pt}}}
    \hline
    \multicolumn{1}{c|}{Layer} &
    \multicolumn{1}{c}{All} &
    \multicolumn{1}{c}{00-01} &
    \multicolumn{1}{c}{02-03} &
    \multicolumn{1}{c}{04-05} &
    \multicolumn{1}{c}{06-07} &
    \multicolumn{1}{c}{08-17}                         \\\hline
    Pose    & 0.51 & 0.42 & 0.20 & 0.03 & 0.01 & 0.00 \\
    Smile   & 0.50 & 0.02 & 0.32 & 0.24 & 0.08 & 0.01 \\
    Age     & 0.54 & 0.09 & 0.20 & 0.23 & 0.19 & 0.04 \\
    Gender  & 0.45 & 0.05 & 0.44 & 0.10 & 0.02 & 0.00 \\
    Glasses & 0.41 & 0.23 & 0.28 & 0.01 & 0.00 & 0.00 \\\hline
  \end{tabular}
  \vspace{-5pt}
\end{table}

\setlength{\tabcolsep}{10pt}
\begin{table*}[t]
  \caption{
    \textbf{Identity discrepancy} after the face manipulation using InterFaceGAN.
    Larger number (within range [0, 1]) means lower similarity.
  }
  \label{tab:identity}
  \vspace{-8pt}
  \centering\small
  \begin{tabular}{c|*{1}{C{50pt}}|*{1}{C{50pt}}|*{6}{C{30pt}}}
    \hline
    \multicolumn{1}{c|}{} &
    \multicolumn{1}{c|}{PGGAN} &
    \multicolumn{1}{c|}{StyleGAN} &
    \multicolumn{6}{c}{StyleGAN $\W$ Space}                         \\
    \multicolumn{1}{c|}{Layer} &
    \multicolumn{1}{c|}{$\Z$ Space} &
    \multicolumn{1}{c|}{$\Z$ Space} &
    \multicolumn{1}{c}{All} &
    \multicolumn{1}{c}{00-01} &
    \multicolumn{1}{c}{02-03} &
    \multicolumn{1}{c}{04-05} &
    \multicolumn{1}{c}{06-07} &
    \multicolumn{1}{c}{08-17}                                       \\\hline
    Pose    & 0.48 & 0.41 & 0.46 & 0.39 & 0.28 & 0.07 & 0.03 & 0.01 \\
    Smile   & 0.24 & 0.31 & 0.21 & 0.04 & 0.20 & 0.10 & 0.04 & 0.01 \\
    Age     & 0.53 & 0.47 & 0.28 & 0.12 & 0.18 & 0.09 & 0.06 & 0.01 \\
    Gender  & 0.61 & 0.51 & 0.40 & 0.11 & 0.37 & 0.13 & 0.03 & 0.01 \\
    Glasses & 0.55 & 0.49 & 0.37 & 0.21 & 0.29 & 0.09 & 0.06 & 0.01 \\\hline
  \end{tabular}
  \vspace{-5pt}
\end{table*}

Table~\ref{tab:layerwise} and Figure~\ref{fig:layerwise} show the quantitative and qualitative results respectively.
From Table~\ref{tab:layerwise}, we can see that ``pose'' is mostly controlled at layer 00-01, ``smile'' is controlled at layer 02-05, ``age'' is controlled at layer 02-07, ``gender'' is controlled at layer 02-03, and ``eyeglasses'' is controlled at layer 00-03.
All attributes are barely affected by editing layer 08-17.
That is because layer 08-17 mainly correspond to texture, such as skin color and background~\cite{stylegan}.
We may identify their roles if we have texture-related attribute classifiers.
Visualization results in Figure~\ref{fig:layerwise} also gives the same conclusion.
It implies that GANs learn different representations at different layers.
This provides us with some insights into a better understanding of the learning mechanism of GANs.

\subsection{Effect of Learned Semantics on Face Identity}\label{subsec:identity-analysis}
Identity is essential for face analysis, thus we perform the identity analysis to see how identity information varies in the manipulation process using InterFaceGAN.
We employ a face recognition model to extract the identity features from the faces before and after semantic editing.
The face recognition model is trained for the face verification task.
It extracts a $256d$ feature vector from a given face image and uses cosine distance as the metric to evaluate the similarity between two subjects.

Table~\ref{tab:identity} shows the results corresponding to different latent spaces from different models, from which we have the following observations:
(i) ``Gender'' affects the identity most and ``smile'' affects the identity least. To some extent, this can be used to verify how sensitive the face identity is to a particular attribute. For example, ``pose'' and ``eyeglasses'' seem to also affect the identity a lot. This makes sense since large pose is always the obstacle in face recognition task and eyeglasses are commonly used to disguise identity in the real world. We may use InterFaceGAN to synthesize more hard samples to in turn improve the face recognition model.
(ii) StyleGAN $\W$ space best preserves the identity information due to its disentanglement property. That is because identity is much more complex than other semantics. A more disentangled representation is helpful in identity control.
(iii) As for the layer-wise results, we can get a similar conclusion to the layer-wise analysis in Section~\ref{subsec:layerwise-analysis}.

\begin{figure*}[t]
  \centering
  \includegraphics[width=1.0\linewidth]{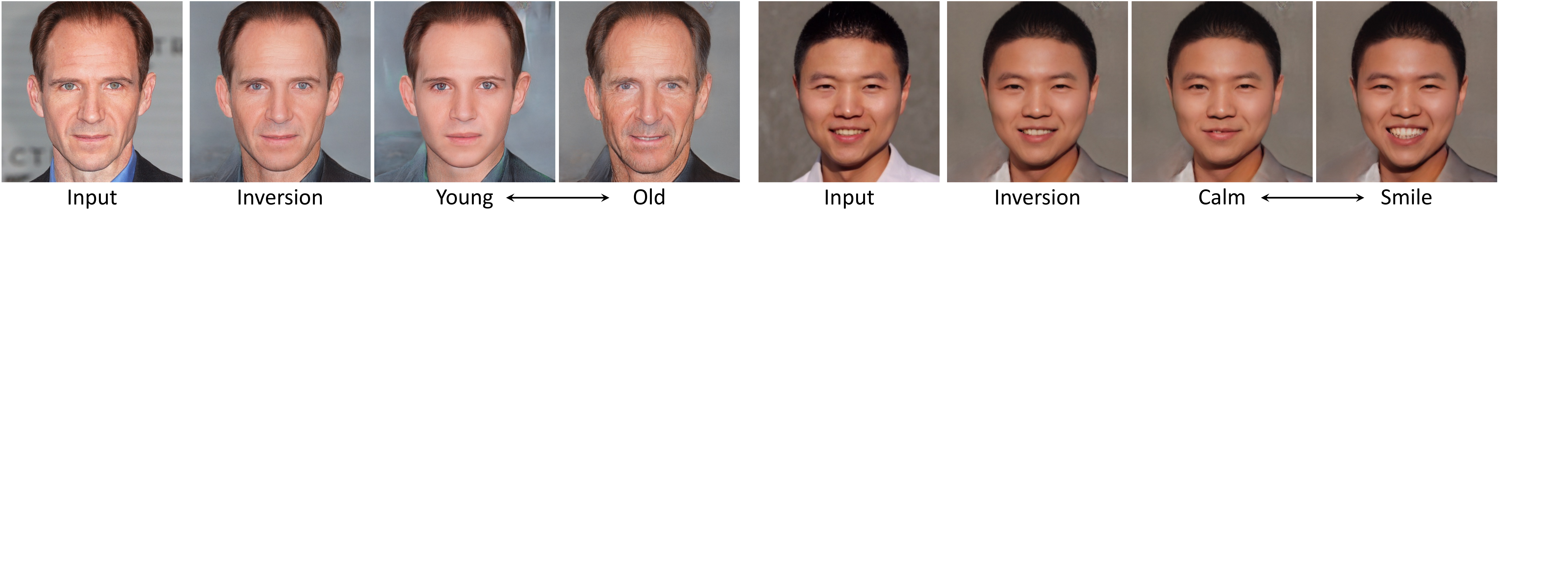}
  \vspace{-22pt}
  \caption{
    \textbf{Manipulating real faces with GAN inversion}.
    Given a target image, we first invert it by optimizing the latent code and then manipulate the inverted code with InterFaceGAN.
    StyleGAN \cite{stylegan} is used as the generative model.
  }
  \label{fig:real-image-manipulation-inversion}
  \vspace{-5pt}
\end{figure*}

\begin{figure*}[t]
  \centering
  \includegraphics[width=1.0\linewidth]{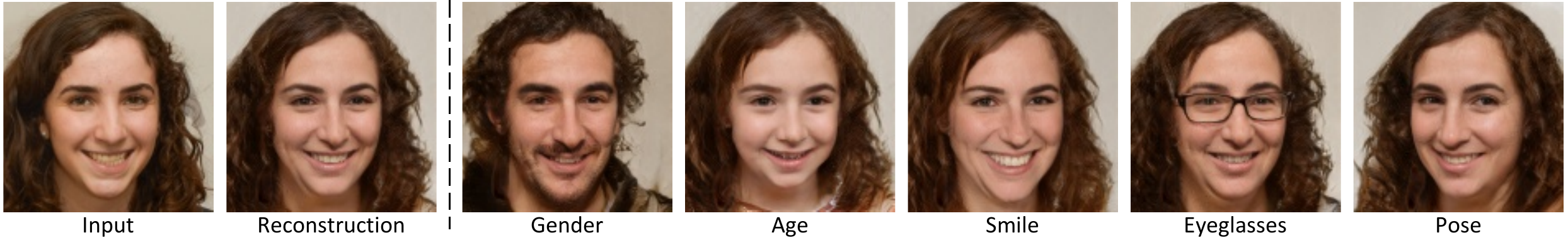}
  \vspace{-22pt}
  \caption{
    \textbf{Manipulating real faces with encoder-decoder generative model}, LIA \cite{lia}.
  }
  \label{fig:real-image-manipulation-encoder}
  \vspace{-5pt}
\end{figure*}

\begin{figure*}[t]
  \centering
  \includegraphics[width=1.0\linewidth]{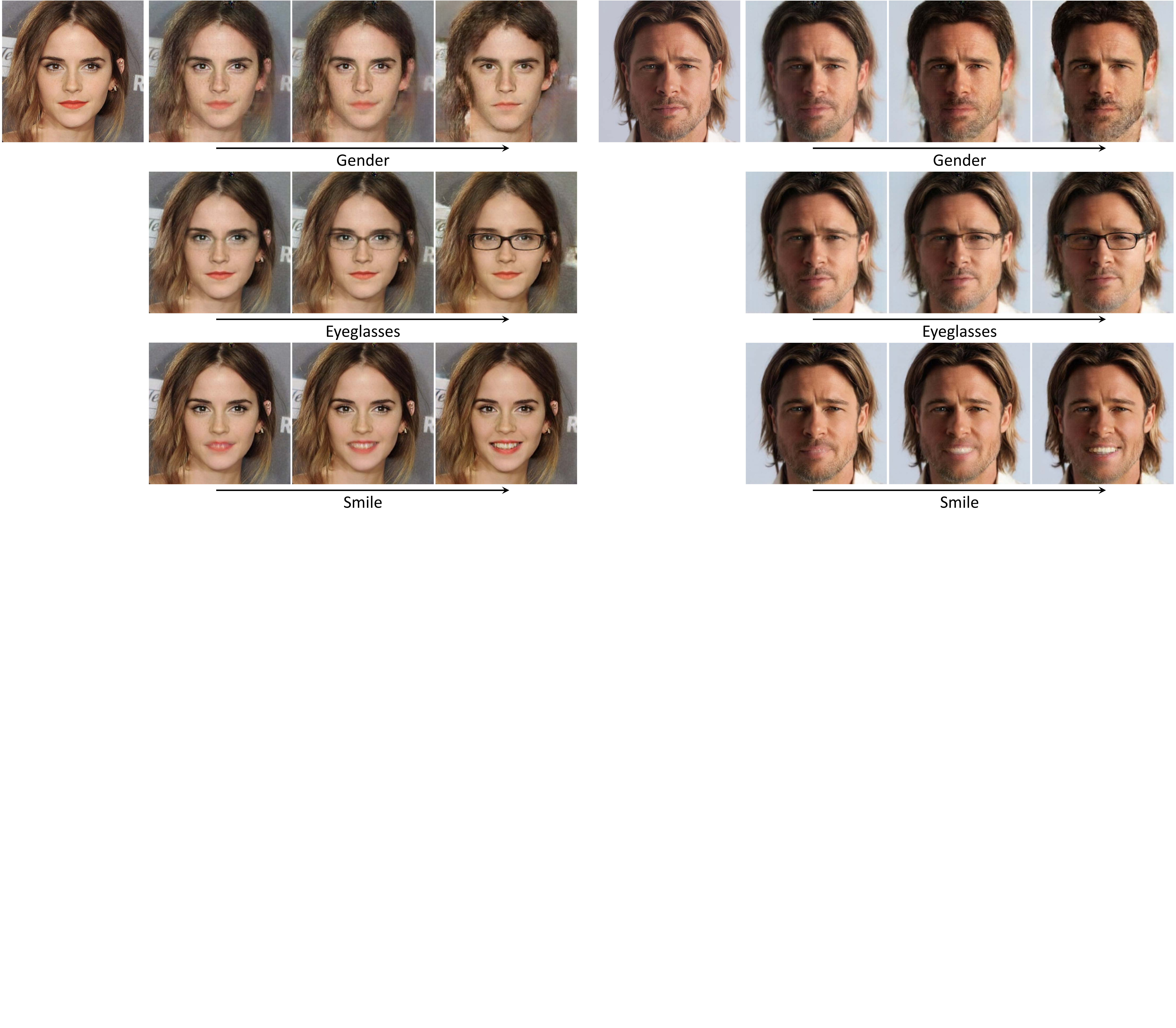}
  \vspace{-22pt}
  \caption{
    \textbf{Real image manipulation by learning additional pix2pixHD models} \cite{pix2pixhd} on synthetic dataset collected from InterFaceGAN.
    DNI \cite{wang2019deep} is used to get continuously varying results.
    %
  }
  \label{fig:real-image-manipulation-learn}
  \vspace{-5pt}
\end{figure*}

\section{Real Image Manipulation}\label{sec:real-image-manipulation}
In this section, we apply the semantics implicitly learned by GANs to real face editing.
Since most of the GAN models lack the inference function to encode real images, we try two approaches.
One is based on GAN inversion, which inverts the target image to the latent code for further editing.
The other uses InterFaceGAN to generate synthetic image pairs and trains additional feed-forward pixel-to-pixel models.

\subsection{Combining GAN Inversion with InterFaceGAN}\label{subsec:inversion-based}

Recall that InterFaceGAN achieves semantic face editing by moving the latent code along a particular direction in the latent space.
Accordingly, for real image editing, one straightforward way is to invert the target face back to a latent code.
It turns out to be a non-trivial task because GANs do not fully capture all the modes and the diversity of the true distribution, which means it is difficult to perfectly recover any real image with a finitely dimensional latent code.
To invert a pre-trained GAN model, there are two typical approaches.
One is the optimization-based approach, which directly optimizes the latent code with the fixed generator to minimize the pixel-wise reconstruction error \cite{image2stylegan,zhu2020indomain}.
The other is encoder-based, where an extra encoder network is trained to learn the inverse mapping \cite{zhu2016generative,lia}.
We integrate these two baseline approaches into InterFaceGAN to see how our proposed manipulation pipeline performs on real image editing.

Figure~\ref{fig:real-image-manipulation-inversion} and Figure~\ref{fig:real-image-manipulation-encoder} show the manipulation results by using optimized-based inversion approach and encoder-based inversion approach respectively.
In Figure~\ref{fig:real-image-manipulation-inversion}, we use StyleGAN \cite{stylegan} as the generative model.
Following prior work \cite{image2stylegan,image2stylegan++,zhu2020indomain}, we treat the layer-wise styles (\emph{i.e.}, $\w$ for all layers) as the optimization target such that the input image can be better reconstructed.
In Figure~\ref{fig:real-image-manipulation-encoder}, we use LIA \cite{lia} as the generative model, which learns an additional encoder beyond the original two-player game in GANs.
We observe that both approaches work well together with InterFaceGAN and we are able to realistically edit the input faces with various attributes, like age and face pose.
It demonstrates the generalization ability of InterFaceGAN.
We also notice that the optimization-based method better preserves the identity information and the face details.
But the encoder-based approach is much faster.
In particular, to invert a single image, the optimization-based method needs around 3 minutes while the encoder-based method needs less than 0.1 seconds.
More results can be found in \textbf{Appendix}.

\subsection{Training with Paired Synthetic Data Collected from InterFaceGAN}\label{subsec:learning-based}

Another way to apply InterFaceGAN to real image editing is to train additional models.
Unlike existing face manipulation models \cite{faceidgan,lample2017fader,stargan} that are trained on a real dataset, we use InterFaceGAN to build a synthetic dataset for training.
There are two advantages:
(i) With recent advancement, GANs can produce high-quality image \cite{stylegan,stylegan2}, significantly narrowing the domain gap.
(ii) With the strong manipulation capability of InterFaceGAN, we can easily create unlimited paired data, which is difficult to obtain in the real world. Taking eyeglasses editing as an example, we can sample numerous latent codes, move them along the ``eyeglasses'' direction, and re-score them to select the ones with highest score change.
With paired data as input and supervision, we train pix2pixHD \cite{pix2pixhd} to achieve face editing.
However, pix2pixHD has its shortcoming that we cannot manipulate the attribute in a continuous manner~\cite{viazovetskyi2020stylegan2}.
To solve this problem, we introduce DNI \cite{wang2019deep} by first training an identical mapping network and then fine-tuning it for a particular attribute.
We summarize the training pipeline as:
\begin{enumerate}
  \vspace{-3pt}
  \item Prepare $10K$ synthetic pairs for each attribute.
  \item Learn an identical pix2pixHD model, which is to transfer the input domain to itself.
  \item Fine-tune the model to transfer a certain attribute by using the original synthesis as the input and the manipulation results with InterFaceGAN as the supervision.
  \item Interpolate the model weights from the identical model to the fine-tuned model for gradual editing.
\end{enumerate}

We choose ``gender'', ``eyeglasses'', and ``smile'' as the target attributes.
Figure~\ref{fig:real-image-manipulation-learn} visualizes some results.
We can conclude that:
(i) The pix2pixHD models trained on the synthetic dataset can successfully manipulate the input face with respect to the target attribute. It suggests that the data generated by InterFaceGAN can well support model training, which may lead to more applications.
(ii) For ``gender'' attribute, we only use female as the input and use male as the supervision. However, after the model training, we can even use this model to add moustache onto male faces as shown in Figure~\ref{fig:real-image-manipulation-learn} (right example).
(iii) Following DNI \cite{wang2019deep}, we interpolate the weights between the identical model and the fine-tuned model. By doing so, we can gradually manipulate the attributes of the input face. The main advantage of learning an additional feed-forward model is its fast inference speed. Also, compared to the encoder-based inversion approach in Section~\ref{subsec:inversion-based}, pix2pixHD model better preserves the identity information.
(iv) During the weight interpolation process, we find that ``smile'' attribute does not perform as well as ``gender'' and ``eyeglasses''. That is because smiling is not a simple pixel-to-pixel translation task but requires a reasonable movement of lips. It is also why pix2pixHD can not be applicable to learning pose rotation, which requires larger movement. Accordingly, the primary limitation of this kind of approach is that it can only transfer some easy-to-map semantics, such as wearing ``eyeglasses''.

\section{Discussion and Conclusion}\label{sec:conclusion}
Interpreting the representation learned by GANs and the disentanglement of attributes is vital for understanding the internal mechanism of deep generative models.
In this work, we provide some pilot studies in this direction by taking face synthesis models as an example.
There are many future works to be done.
As our visual world is far more complicated than faces, it will be interesting to look into the generative models trained to synthesize generic objects and scenes.
For example, for scene generation, besides learning semantics for the entire image, the model should create a spatial layout and learn to synthesize any individual objects inside.
From this perspective, we need a more general method to interpret the GAN models beyond faces.
For face models, there are also many directions worth further studying.
As we have already discussed in Section~\ref{subsec:latent-space-manipulation}, our method may fail for long-distance manipulation due to the linear assumption.
More adaptive and expressive manipulation models, such as non-linear models, would solve this problem
On the other hand, we use off-the-shelf classifiers as auxiliary predictors.
This limits the semantics we can find in the latent space since we may not have the proper classifiers or the attribute may not be well defined or annotated.
Hence, identifying the semantics emerging from synthesizing images in an unsupervised learning manner is worth further exploring.

To conclude, we interpret the disentangled face representation learned by GANs and conduct a thorough study on the emerging facial semantics.
By leveraging the semantic knowledge encoded in the latent space, we are able to edit the attributes in face images realistically.
The conditional manipulation technique is further introduced to disentangle different attributes for more precise control of face editing.
Further experiments suggest that InterFaceGAN is applicable to real image manipulation.

\ifCLASSOPTIONcompsoc
  \section*{Acknowledgments}
\else
  \section*{Acknowledgment}
\fi
This work is supported in part by the Early Career Scheme (ECS) through the Research Grants Council of Hong Kong under Grant No.24206219, in part by RSFS grant from CUHK Faculty of Engineering, and in part by SenseTime Collaborative Grant.

\ifCLASSOPTIONcaptionsoff
  \newpage
\fi
\bibliographystyle{IEEEtran}
\bibliography{ref}

\appendices

\section{Proof}\label{appendix:proof}
In this part, we provide detailed proof of \emph{\textbf{Property 2}} in the main paper.
Recall this property as follows.

\vspace{2pt}\emph{\textbf{Property 2}
Given $\n\in\R^d$ with $\n^T\n=1$, which defines a hyperplane, and a multivariate random variable $\z\sim\Norm(\0,\I_d)$, we have $\Prob(|\n^T\z|\leq2\alpha~\sqrt{\frac{d}{d-2}})\geq(1-3e^{-c d})(1-\frac{2}{\alpha}e^{-\alpha^2/2})$ for any $\alpha\geq1$ and $d\geq4$. Here $\Prob(\cdot)$ stands for probability and $c$ is a fixed positive constant.
}

\vspace{5pt}\noindent\emph{Proof.}

Without loss of generality, we fix $\n$ to be the first coordinate vector.
Accordingly, it suffices to prove that $\Prob(|z_1|\leq2\alpha~\sqrt{\frac{d}{d-2}})\geq(1-3e^{-c d})(1-\frac{2}{\alpha}e^{-\alpha^2/2})$, where $z_1$ denotes the first entry of $\z$.

\begin{figure}[t]
  \centering
  \includegraphics[width=0.85\linewidth]{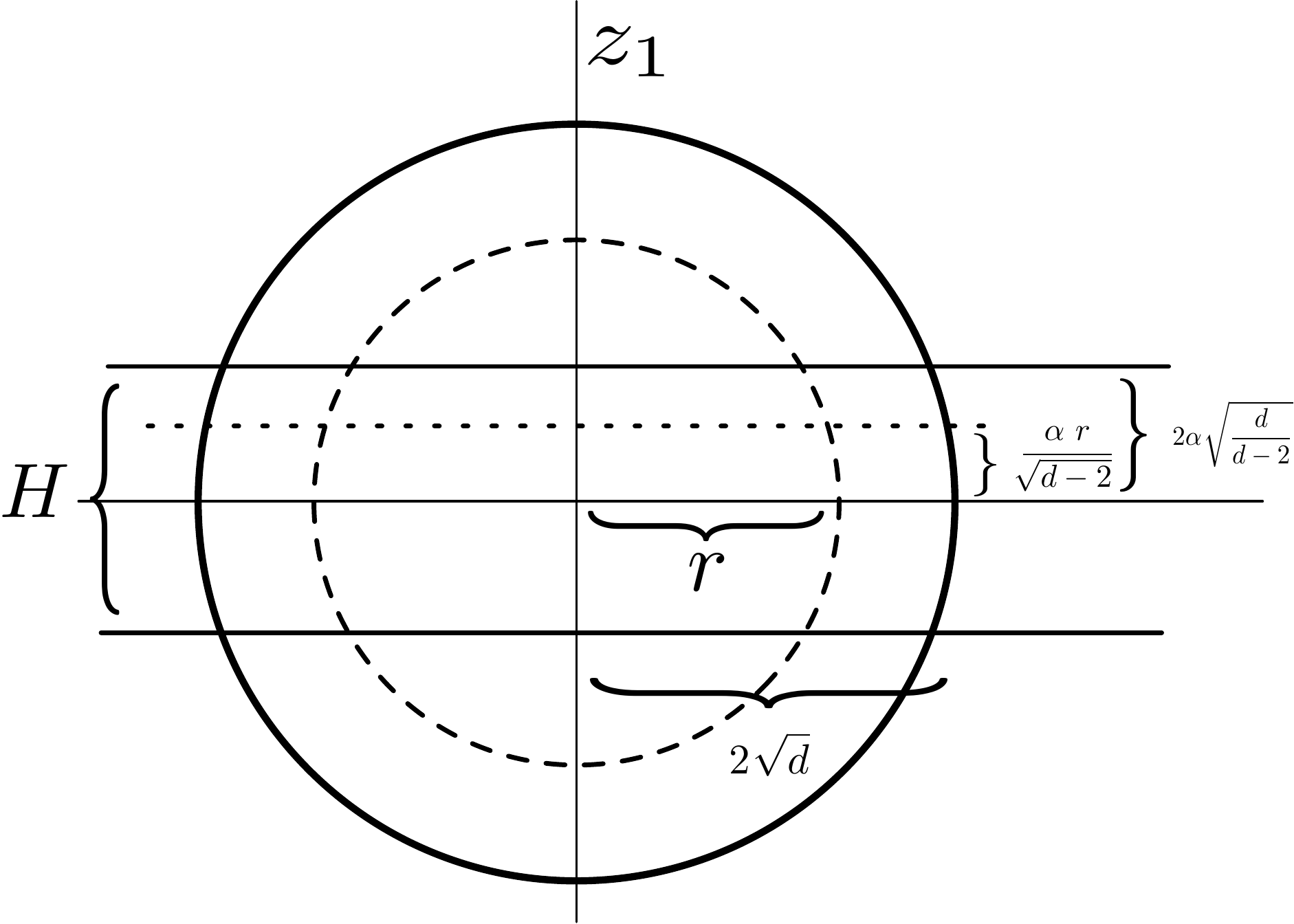}
  \vspace{-5pt}
  \caption{
    Illustration of \emph{\textbf{Property 2}}, which shows that most of the probability mass of high-dimensional Gaussian distribution lies in the thin slab near the ``equator''.
  }
  \vspace{-5pt}
  \label{appendix:fig:property-2}
\end{figure}

As shown in Figure~\ref{appendix:fig:property-2}, let $H$ denote the set
\begin{align}
\{\z\sim\N(\0,\I_d):||\z||_2\leq2\sqrt{d}, |z_1|\leq2\alpha\sqrt{\frac{d}{d-2}}\}, \nonumber
\end{align}
where $||\cdot||_2$ stands for the $l_2$ norm.
Obviously, we have $\Prob(H)\leq\Prob(|z_1|\leq2\alpha\sqrt{\frac{d}{d-2}})$.
Now, we will show $\Prob(H)\geq(1-3e^{-c d})(1-\frac{2}{\alpha}e^{-\alpha^2/2})$

Considering the random variable $R=||\z||_2$, with cumulative distribution function $F(R\leq r)$ and density function $f(r)$, we have
\begin{align}
  \Prob(H) &= \Prob(|z_1|\leq2\alpha\sqrt{\frac{d}{d-2}}|R\leq2\sqrt{d})\Prob(R\leq2\sqrt{d}) \nonumber \\
           &= \int_0^{2\sqrt{d}}\Prob(|z_1|\leq2\alpha\sqrt{\frac{d}{d-2}}|R=r)f(r)dr. \nonumber
\end{align}

According to \emph{Theorem 1} below, when $r\leq2\sqrt{d}$, we have
\begin{align}
   \Prob(H) &= \int_0^{2\sqrt{d}}\Prob(|z_1|\leq2\alpha\sqrt{\frac{d}{d-2}}|R=r)f(r)dr \nonumber \\
            &= \int_0^{2\sqrt{d}}\Prob(|z_1|\leq\frac{2\sqrt{d}}{r}\frac{\alpha}{\sqrt{d-2}}|R=1)f(r)dr \nonumber \\
            &\geq \int_0^{2\sqrt{d}}\Prob(|z_1|\leq\frac{\alpha}{\sqrt{d-2}}|R=1)f(r)dr \nonumber \\
            &\geq \int_0^{2\sqrt{d}}(1-\frac{2}{\alpha}e^{-\alpha^2/2})f(r)dr \nonumber \\
            &= (1-\frac{2}{\alpha}e^{-\alpha^2/2})\int_0^{2\sqrt{d}}f(r)dr \nonumber \\
            &= (1-\frac{2}{\alpha}e^{-\alpha^2/2})\Prob(0\leq R\leq2\sqrt{d}). \nonumber
\end{align}

Then, according to \emph{Theorem 2} below, by setting $\beta=\sqrt{d}$, we have
\begin{align}
  \Prob(H) &= (1-\frac{2}{\alpha}e^{-\alpha^2/2})\Prob(0\leq R\leq2\sqrt{d}) \nonumber \\
           &\geq (1-\frac{2}{\alpha}e^{-\alpha^2/2})(1-3e^{-c d}). \nonumber
\end{align}

Q.E.D.

\begin{figure}[t]
  \centering
  \includegraphics[width=0.70\linewidth]{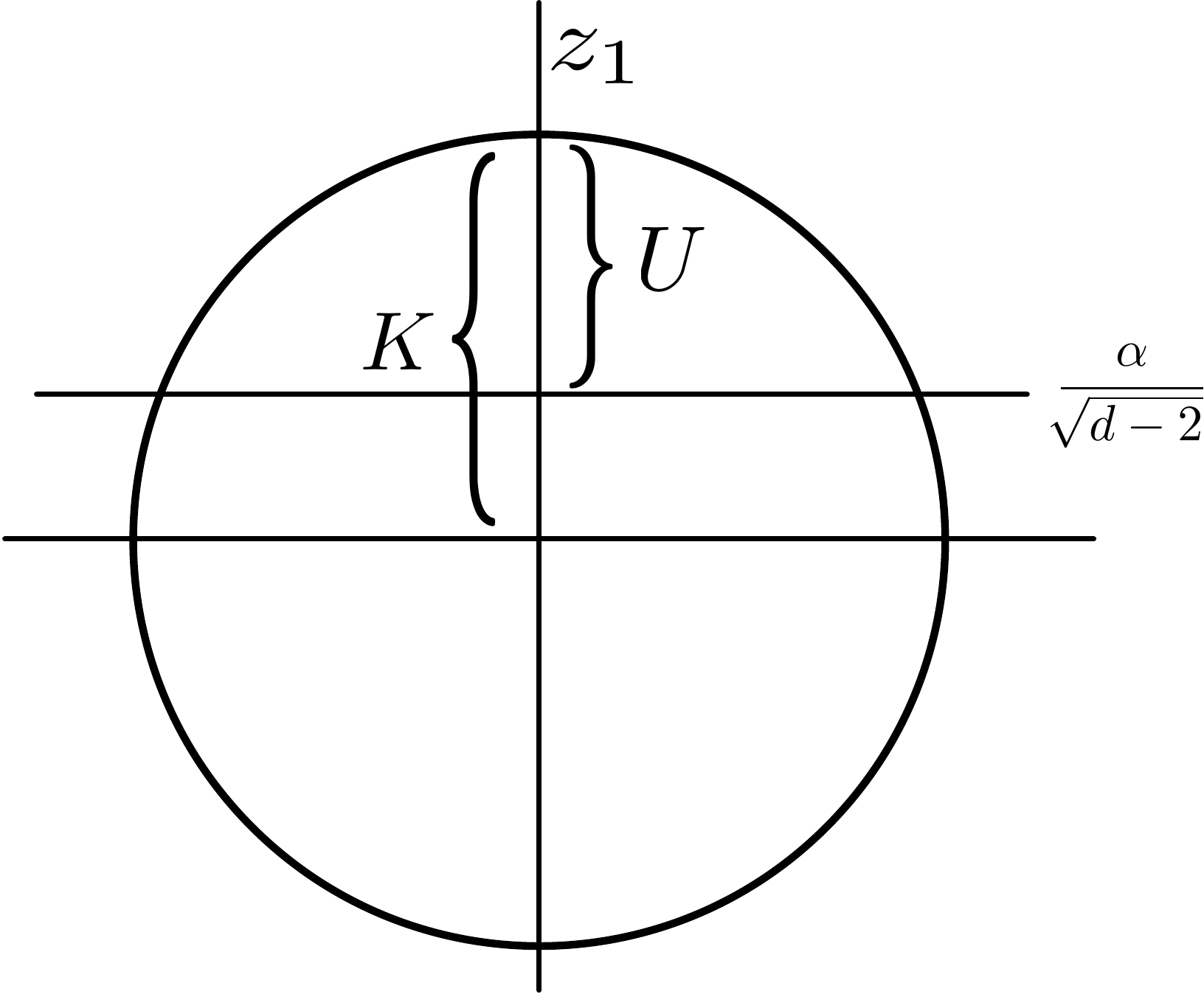}
  \vspace{-5pt}
  \caption{
    Diagram for \emph{\textbf{Theorem 1}}.
  }
  \label{appendix:fig:theorem-1}
  \vspace{-5pt}
\end{figure}

\vspace{2pt}\emph{\textbf{Theorem 1}
Given a unit spherical $\{\z\in\R^d:||\z||_2=1\}$, we have $\Prob(|z_1|\leq\frac{\alpha}{\sqrt{d-2}})\geq1-\frac{2}{\alpha}e^{-\alpha^2/2}$ for any $\alpha\geq1$ and $d\geq4$.}

\vspace{5pt}\noindent\emph{Proof.}

By symmetry, we just prove the case where $z_1\geq0$.
Also, we only consider about the case where $\frac{\alpha}{\sqrt{d-2}} \leq 1$.

Let $U$ denote the set $\{\z\in\R^d:||\z||_2=1,z_1\geq\frac{\alpha}{\sqrt{d-2}}\}$, and $K$ denote the set $\{\z\in\R^d:||\z||_2=1,z_1\geq0\}$.
It suffices to prove that the surface of $U$ area and the surface of $K$ area in Figure~\ref{appendix:fig:theorem-1} satisfy
\begin{align}
  \frac{surf(U)}{surf(K)}\leq\frac{2}{\alpha}e^{-\alpha^2/2}, \nonumber
\end{align}
where $surf(\cdot)$ stands for the surface area of a high dimensional geometry.
Let $A(d)$ denote the surface area of a $d$-dimensional unit-radius ball. Then, we have
\begin{align}
  surf(U) &=\int_{\frac{\alpha}{\sqrt{d-2}}}^{1} (1-z_1^2)^{\frac{d-2}{2}}A(d-1)dz_1 \nonumber \\
          &\leq \int_{\frac{\alpha}{\sqrt{d-2}}}^{1} e^{-\frac{d-2}{2}z_1^2}A(d-1)dz_1 \nonumber \\
          &\leq \int_{\frac{\alpha}{\sqrt{d-2}}}^{1} \frac{z_1\sqrt{d-2}}{\alpha}e^{-\frac{d-2}{2}z_1^2}A(d-1)dz_1 \nonumber \\
          &\leq \int_{\frac{\alpha}{\sqrt{d-2}}}^{\infty} \frac{z_1\sqrt{d-2}}{\alpha}e^{-\frac{d-2}{2}z_1^2}A(d-1)dz_1 \nonumber \\
          &= \frac{A(d-1)}{\alpha\sqrt{d-2}}e^{-\alpha^2/2}. \nonumber
\end{align}

Similarly, we have
\begin{align}
  surf(K) &=\int_{0}^{1} (1-z_1^2)^{\frac{d-2}{2}}A(d-1)dz_1 \nonumber \\
          &\geq \int_{0}^{\frac{1}{\sqrt{d-2}}} (1-z_1^2)^{\frac{d-2}{2}}A(d-1)dz_1 \nonumber \\
          &\geq \frac{1}{\sqrt{d-2}} (1-\frac{1}{d-2})^{\frac{d-2}{2}}A(d-1). \nonumber
\end{align}

Considering the fact that $(1-x)^a\geq1-ax$ for any $a\geq1$ and $0\leq x\leq1$, we have
\begin{align}
  surf(K) &\geq \frac{1}{\sqrt{d-2}} (1-\frac{1}{d-2})^{\frac{d-2}{2}}A(d-1) \nonumber \\
          &\geq \frac{1}{\sqrt{d-2}} (1-\frac{1}{d-2}\frac{d-2}{2})A(d-1) \nonumber \\
          &= \frac{A(d-1)}{2\sqrt{d-2}}. \nonumber
\end{align}

Accordingly,
\begin{align}
  \frac{surf(U)}{surf(K)} \leq \frac{\frac{A(d-1)}{\alpha\sqrt{d-2}}e^{-\alpha^2/2}}{\frac{A(d-1)}{2\sqrt{d-2}}}
                           = \frac{2}{\alpha}e^{-\alpha^2/2}. \nonumber
\end{align}

Q.E.D.

\vspace{2pt}\emph{\textbf{Theorem 2 (Gaussian Annulus Theorem {\rm \cite{blum2020foundations}})}
For a $d$-dimensional spherical Gaussian with unit variance in each direction, for any $\beta\leq\sqrt{d}$, all but at most $3e^{-c\beta^2}$ of the probability mass lies within the annulus $\sqrt{d}-\beta\leq||\z||_2\leq\sqrt{d}+\beta$, where $c$ is a fixed positive constant.
}

That is to say, given $\z\sim\N(\0, \I_d)$, $\beta\leq\sqrt{d}$, and a constant $c>0$, we have
\begin{align}
  \Prob(\sqrt{d}-\beta\leq||\z||_2\leq\sqrt{d}+\beta)\geq(1-3e^{-c\beta^2}). \nonumber
\end{align}

\begin{figure*}[t]
  \centering
  \includegraphics[width=1.0\linewidth]{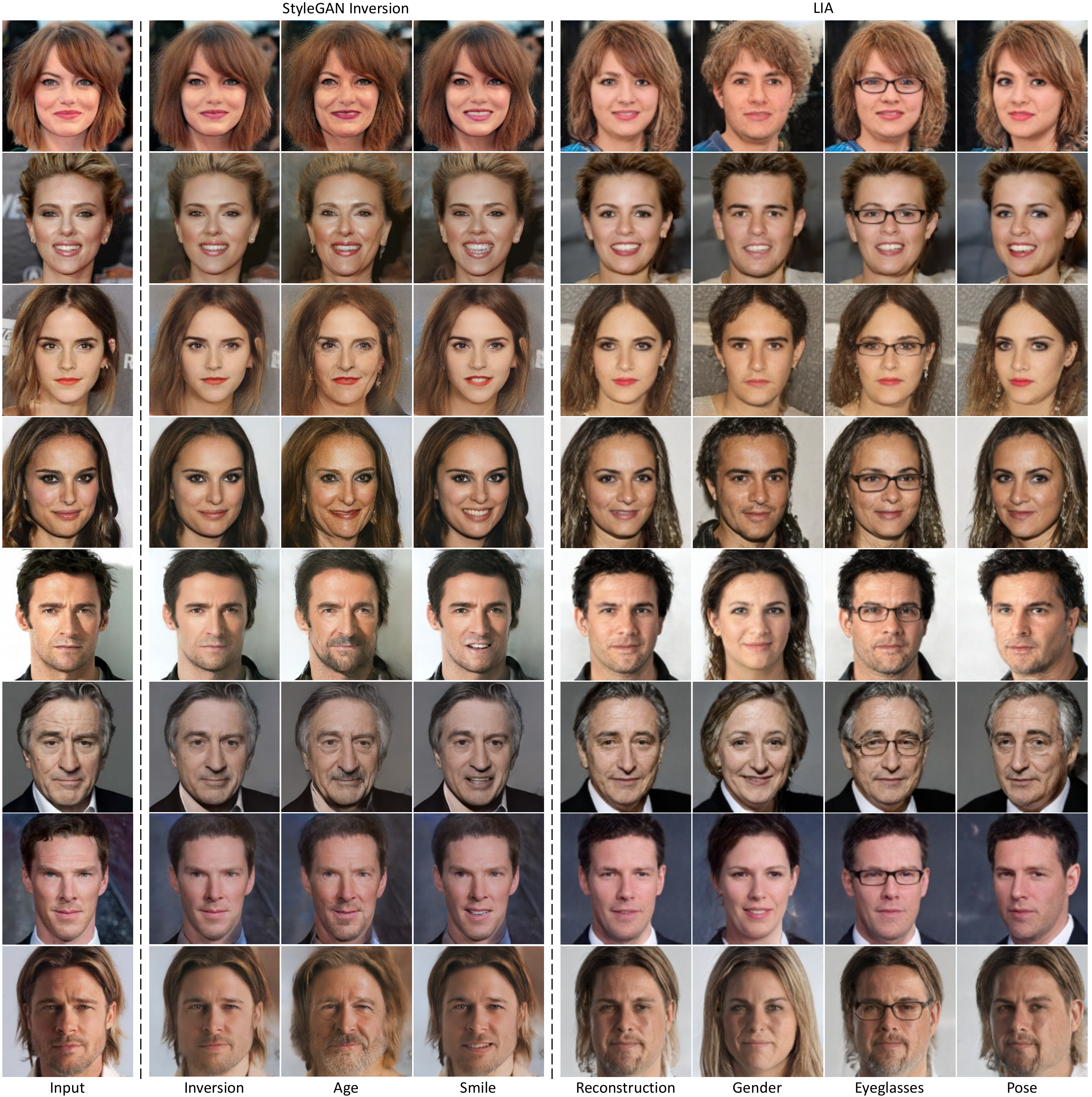}
  \vspace{-22pt}
  \caption{
    \textbf{Real image manipulation} with optimization-based GAN inversion approach \cite{zhu2020indomain}, and encoder-based GAN inversion approach \cite{lia}.
  }
  \label{appendix:fig:real-image-manipulation}
  \vspace{0pt}
\end{figure*}

\section{Real Image Manipulation}\label{appendix:real-image-editing}
Besides the examples shown in the main paper, we present more results on real face editing by integrating InterFaceGAN with GAN inversion methods, including both optimization-based \cite{zhu2020indomain} and encoder-based \cite{lia}.
Figure~\ref{appendix:fig:real-image-manipulation} compares these two approaches.
We can tell that the optimization-based method better recovers the input images and hence better preserves the identity information.
But for both methods, the interpretable semantics inside the latent representation are capable of faithfully editing the corresponding facial attributes of the reconstructed face.

\end{document}